\PassOptionsToPackage{table,xcdraw}{xcolor}
\documentclass[manuscript,screen]{acmart}
\usepackage{amsmath,amsfonts}
\usepackage{array}
\usepackage[caption=false,font=normalsize,labelfont=sf,textfont=sf]{subfig}
\usepackage{textcomp}
\usepackage{stfloats}
\usepackage{url}
\usepackage{verbatim}
\usepackage{graphicx}
\usepackage[ruled,vlined,linesnumbered,noresetcount,algo2e]{algorithm2e}
\usepackage{algorithm}
\usepackage{wrapfig}
\usepackage[acronym,toc]{glossaries}
\usepackage{tablefootnote}
\usepackage{tabularx}
\usepackage{float}
\restylefloat{table}
\usepackage{longtable}
\usepackage{multirow}
\usepackage{wrapfig}
\usepackage{caption}
\usepackage[switch]{lineno}
\usepackage{algorithmic}
\usepackage[ruled]{algorithm2e} 
\usepackage{xcolor} 

\def\HiLi{\leavevmode\rlap{\hbox to \hsize{\color{gray!50}\leaders\hrule height .8\baselineskip depth .5ex\hfill}}}

\usepackage{soul}
\def\BibTeX{{\rm B\kern-.05em{\sc i\kern-.025em b}\kern-.08em
 T\kern-.1667em\lower.7ex\hbox{E}\kern-.125emX}}
\usepackage{balance}

\SetKwInput{KwConst}{Constants}
\SetKwInput{KwAlg}{Main Algorithm}
\SetKwInput{KwCircuits}{Parallel Circuits}
\usepackage{lineno}
\definecolor{maroon}{cmyk}{0,0.87,0.68,0.32}

\usepackage{orcidlink}
\newacronym{mhsnn}{MHSNN}{multi-hierarchical spiking neural network}
\newacronym{resume}{ReSuMe}{Remote Supervised Method}
\newacronym{stdp}{STDP}{Spike Timing Dependent Plasticity}
 
\newacronym{hsmd}{HSMD}{Hybrid Sensitive Motion Detector}
\newacronym{neurohsmd}{NeuroHSMD}{Neuromorphic Hybrid Sensitive Motion Detector}
\newacronym{ijcnn}{IJCNN}{International Joint Conference on Neural Networks}

\newacronym{fps}{fps}{Frames Per Second}

\newacronym{cdnet2012}{CDnet2012}{2012 Change Detection}
\newacronym{cdnet2014}{CDnet2014}{2014 Change Detection}

\newacronym{adas}{ADAS}{Advanced-Driver Assist Systems}
\newacronym{ns}{NS}{Navigational Systems}
\newacronym{cots}{COTS}{Commercial-Off-The-Shelf}

\newacronym{bs}{BS}{Background subtraction}
\newacronym{dbs}{DBS}{Dynamic Background subtraction}
\newacronym{roi}{ROI}{Region of Interest}
\newacronym{ptz}{PTZ}{Pan, Tilt and Zoom}

\newacronym{tp}{TP}{True Positives}
\newacronym{tn}{TN}{True Negatives}
\newacronym{fp}{FP}{False Positives}
\newacronym{fn}{FN}{False Negatives}
\newacronym{nmet}{nMet}{Number of Metrics}
\newacronym{pcc}{PCC}{Percentage of Correct Classifications}
\newacronym{pwc}{PWC}{Percentage of Wrong Classifications}

\newacronym{re}{Re}{Recall}
\newacronym{sp}{Sp}{Specificity}
\newacronym{fpr}{FPR}{False Positive Rate}
\newacronym{fnr}{FNR}{False Negative Rate}
\newacronym{wcr}{WCR}{Wrong Classifications Rate}
\newacronym{ccr}{CCR}{Correct Classifications Rate}
\newacronym{pr}{Pr}{Precision}
\newacronym{f1}{F1}{F-measure}
\newacronym{r}{R}{Average Ranking}
\newacronym{arc}{RC}{Average Ranking across all Categories}

\newacronym{ml}{ML}{Machine Learning}
\newacronym{rf}{RF}{Receptive Field}
\newacronym{hmi}{HMI}{Human Machine Interaction}
\newacronym{pca}{PCA}{Principal Component Analysis}
\newacronym{ica}{ICA}{Independent Component Analysis}
\newacronym{rpca}{RPCA}{Robust Principal Component Analysis}
\newacronym{ram-nn}{RAM-NN}{Random Access Memory Neural Network}
\newacronym{ram}{RAM}{Random Access Memory}
\newacronym{rbf-nn}{RBF-NN}{Radial Basis Function Neural Network}
\newacronym{sobs}{SOBS}{Self Organizing Background Subtraction}
\newacronym{hsv}{HSV}{Hue, Saturation and brightness Value}
\newacronym{dnn}{DNN}{Deep Neural Network}
\newacronym{rbm}{RBM}{Restricted Boltzman Machine}
\newacronym{sdae}{SDAE}{Stacked Denoising Auto-Encoder}
\newacronym{nerem}{NeREM}{Neural Response Mixture}
\newacronym{hog}{HOG}{Histogram of Oriented Gradients}
\newacronym{gloh}{GLOH}{Gradient Location and Orientation Histogram}
\newacronym{surf}{SURF}{Speeded Up Robust Features}
\newacronym{sift}{SIFT}{Scale-Invariant Feature Transform}
\newacronym{brief}{BRIEF}{Binary Robust Independent Elementary Features}
\newacronym{orb}{ORB}{Oriented FAST and Rotated BRIEF}
\newacronym{capoa}{CAPOA}{Content-Adaptive Progressive Occlusion analysis}
\newacronym{dlt}{DLT}{Deep Learning Tracker}
\newacronym{codd}{CODD}{Custom object direction detection}
\newacronym{stl}{STL}{Standard Template Library}

\newacronym{dsgc}{DSGC}{Direction-Selective Ganglion Cells}
\newacronym{ann}{ANN}{Artificial Neural Network}
\newacronym{rsnn}{RSNN}{Recurrent Spiking Neural Network}
\newacronym{cnn}{CNN}{Convolutional Neural Network}
\newacronym{rbcnn}{RBCNN}{Region Based Convolutional Neural Network}
\newacronym{csnn}{CSNN}{Convolutional Spiking Neural Network}
\newacronym{noc}{NoC}{Network-on-Chip}
\newacronym{rtss}{RTSS}{Real-Time Semantic Segmentation}
\newacronym{sbs}{SBS}{Semantic Background Segmentation}
\newacronym{mfcn}{MFCN}{Multiscale Fully Convolutional Network}

\newacronym{nha}{NHA}{NeuronHSMD Host Application}
\newacronym{ndk}{NDK}{NeuronHSMD device kernels}
\newacronym{sdk}{SDK}{Software Development Kit}

\newacronym{snn}{SNN}{Spiking Neural Network}

\newacronym{gsoc}{GSOC}{Google Summer of Code}
\newacronym{mog}{MOG}{Mixture of Gaussians}
\newacronym{mog2}{MOG2}{Gaussian Mixture Probability Density}
\newacronym{knn}{KNN}{Mixture of Gaussians K Nearest Neighbours}
\newacronym{lsbp}{LSBP}{Local Single Value Decomposition Binary Pattern}
\newacronym{ieee}{IEEE}{Institute of Electrical and Electronics Engineers}

\newacronym{cv}{CV}{Computer Vision}
\newacronym{opencv}{OpenCV}{Open Computer Vision}
\newacronym{opencl}{OpenCL}{Open Computer Language}
\newacronym{hdl}{HDL}{Hardware Description Language}
\newacronym{hls}{HLS}{High Level Synthesis}
\newacronym{rtl}{RTL}{Register Transfer Level}
\newacronym{uav}{UAV}{Unmanned Aerial Vehicle}
\newacronym{ipl}{IPL}{Inner Plexiform Layer}
\newacronym{sac}{SAC}{starburst amacrine cells}

\newacronym{nn}{NN}{Neural Network}
\newacronym{oms-gc}{OMS-GC}{Object Motion Sensitive Ganglion Cells}
\newacronym{omd}{OMD}{Object Motion Detection}
\newacronym{gc}{GC}{Ganglion Cells}
\newacronym{nuc}{NUC}{Next Unit of Computing}

\newacronym{sram}{SRAM}{Static Random Access Memory}
\newacronym{sdram}{SDRAM}{Synchronous Dynamic Random Access Memory}
\newacronym{vhsic}{VHSIC}{Very High Speed Integrated Circuit}
\newacronym{vhdl}{VHDL}{Very High Speed Integrated Circuit Hardware Description Language}
\newacronym{asic}{ASIC}{Application-Specific Integrated Circuits}
\newacronym{vlsi}{VLSI}{Very Large Scale of Integration}
\newacronym{dvs}{DVS}{Dynamic Voltage Scaling}
\newacronym{aer}{AER}{Address Event Representation}
\newacronym{davis}{DAVIS}{Dynamic Active Pixel Vision Sensor}
\newacronym{nna}{NNA}{Neural Network Accelerators}

\newacronym{fpga}{FPGA}{Field-Programmable Gate Array}
\newacronym{gpu}{GPU}{Graphical Processing Unit}
\newacronym{cpu}{CPU}{Central Processing Unit}
\newacronym{ic}{IC}{Integrated Circuit}
\newacronym{os}{OS}{Operating Systems}

\newacronym{cpg}{CPG}{Central Pattern Generator}

\newacronym{lif}{LIF}{Leaky-Integrate-and-Fire}
\newacronym{iaf}{IAF}{Integrate-and-Fire}
\newacronym{hh}{HH}{Hodgkin \& Huxley}
\newacronym{rc}{RC}{Resitance and Capacitance}
\newacronym{izk}{IZK}{Izhikevich}
\newacronym{ai}{AI}{Artificial Intelligence}
\newacronym{dog}{DoG}{Difference-of-Gaussians}
\newacronym{rgb}{RGB}{Red, Green and Blue}
\newacronym{fir}{FIR}{Finite Impulse Response}
\newacronym{iir}{IIR}{Infinite Impulse Response}

\newacronym{lab}{LAB}{logic array blocks}
\newacronym{mlab}{MLAB}{memory logic array blocks}
\newacronym{alut}{ALU}{adaptive look-up-table}
\newacronym{alm}{ALM}{adaptive logic module}
\newacronym{dsp}{DSP}{digital signal processing}

\newacronym{hdd}{HDD}{Hard Disk Drive}
\newacronym{ssd}{SSD}{Solid State Drive}
\newacronym{usb}{USB}{Universal Serial Bus}
\newacronym{sata}{SATA}{Serial Advanced Technology Attachment}
\newacronym{pcie}{PCIe}{Peripheral Component Interconnect Expres}
\newacronym{cuda}{CUDA}{Compute Unified Device Architecture}
\newacronym{ps}{PS}{Processor System}

\newacronym{som}{SOM}{Self Organizing Map}
\newacronym{bsp}{BSP}{Board Support Package}
\newacronym{ode}{ODE}{Ordinary Differential Equation}

\AtBeginDocument{%
  \providecommand\BibTeX{{%
    \normalfont B\kern-0.5em{\scshape i\kern-0.25em b}\kern-0.8em\TeX}}}

\setcopyright{acmcopyright}
\copyrightyear{2022}
\acmYear{2022}
\acmDOI{XXXXXXX.XXXXXXX}

\begin{document}
\title{NeuroHSMD: Neuromorphic Hybrid Spiking Motion Detector} 

\author{Pedro Machado \orcidlink{0000-0003-1760-3871}, Jo\~ao Filipe Ferreira\orcidlink{0000-0002-2510-2412} Andreas Oikonomou\orcidlink{0000-0002-5069-3971}}

\affiliation{
  \institution{Department of Computer Science,
  School of Science and Technology,
  Nottingham Trent University}
  \streetaddress{Clifton Lane}
  \city{Nottingham}
  \country{UK}
  \postcode{NG11 8NS}
}
\email{\{pedro.machado, andreas.oikonomou\}@ntu.ac.uk}

\author{T.M. McGinnity\orcidlink{0000-0002-9897-4748}}
\affiliation{
  \institution{Intelligent Systems Research Centre,
  School of Computing, Engineering and Intelligent Systems,
  Ulster University}
  \streetaddress{Northlands Road, Magee Campus}
  \city{Londonderry}
  \country{UK}
  \postcode{BT48 7JL}
}
\email{tm.mcginnity@ulster.ac.uk}
\renewcommand{\shortauthors}{Machado et al.}

\begin{abstract}
Vertebrate retinas are highly-efficient in processing trivial visual tasks such as detecting moving objects, which still represent complex challenges for modern computers. In vertebrates, the detection of object motion is performed by specialised retinal cells named \gls*{oms-gc}. \gls*{oms-gc} process continuous visual signals and generate spike patterns that are post-processed by the Visual Cortex. Our previous \gls*{hsmd} algorithm was the first hybrid algorithm to enhance \gls*{bs} algorithms with a customised 3-layer \gls*{snn} that generates \gls*{oms-gc} spiking-like responses. In this work, we present a \gls*{neurohsmd} algorithm that accelerates our \gls*{hsmd} algorithm using \glspl*{fpga}. The \gls*{neurohsmd} was compared against the \gls*{hsmd} algorithm, using the same \gls*{cdnet2012} and \gls*{cdnet2014} benchmark datasets. When tested against the \gls*{cdnet2012} and \gls*{cdnet2014} datasets, \gls*{neurohsmd} performs object motion detection at $720\times480$ at 28.06 \gls*{fps} and $720\times480$ at 28.71 \gls*{fps}, respectively, with no degradation of quality. Moreover, the \gls*{neurohsmd} proposed in this paper was completely implemented in \gls*{opencl} and therefore is easily replicated in other devices such as \glspl*{gpu} and clusters of \glspl*{cpu}.
\end{abstract}

\ccsdesc[500]{Computer systems organization~Embedded systems}
\ccsdesc[300]{Computer systems organization~Redundancy}
\ccsdesc{Computer systems organization~Robotics}
\ccsdesc[100]{Networks~Network reliability}

\keywords{ SNN, HMSD, NeuroHSMD, retinal cells, OMS-GC, FPGA, background subtraction, object motion detection}
\maketitle

\section{Introduction}\label{Ch1:intro}

The human brain is characterised by its tolerance to faults/noise, concurrent processing capabilities, flexibility and high level of parallelisation when processing data. Furthermore, the adult human brain has a power consumption of about 400 Kcal per day, equivalent to 25 Watts \cite{Pastur-Romay2017}. Again, the human brain can reach 10-50 petaflops outperforming any \gls*{cots} \acrfull*{cpu} \cite{Brooks2012}. Despite \glspl*{cpu} outperforming the human brain when processing and transmitting sequential signals by several orders of magnitude, the human brain exceeds \glspl*{cpu} processing millions of signals in parallel using its massively parallel circuits \cite{Pastur-Romay2017,Brooks2012}. \glspl*{snn} are well known for their biological plausibility but also by their complexity inherited from biological systems, which are characterised for being massively parallel \cite{Bouvier2019}. Modern computation platforms rely heavily on \glspl*{cpu} to provide compatibility and security with other devices/applications. Although the design \gls*{cpu} paradigm has shifted into multi-core and multi-processor to overcome the limitations associated with the clock speed \cite{zhang2018}, the multi-core and multi-processor strategy will shortly meet technological limitations related to the increase of power consumption of these solutions \cite{zhang2018}.

Machado et al. proposed the \acrfull*{hsmd} algorithm \cite{Machado2021} that has proven to be very sensitive to object motion events as a direct consequence of using an \acrfull*{snn} to emulate the basic functionality observed in \acrfull*{oms-gc} (see Figure~\ref{fig:hsmd}). The \gls*{snn} utilised is composed of 3 layers of neurons interconnected on a 1:1 synaptic connectivity. \glspl*{cpu} which are suitable for sequential operations are not optimised for running massively parallel \glspl*{snn} architectures. Unlike \glspl*{cpu} that are not optimised for parallel tasks, and  \acrfullpl*{gpu} and \acrfullpl*{fpga} are parallel processing devices for accelerating paralellisable algorithms. \glspl*{gpu} are specialised electronic circuits with a flexible architecture designed for parallel processing of graphics, video rendering, and accelerating some types of \gls*{ai} algorithms \cite{ferreira2011,ferreira2012,ferreira2013,Intel2020b}, while \glspl*{fpga} are \glspl*{ic} composed of built-in interconnected hardware blocks that can be freely reprogrammable after manufacturing \cite{Intel2020}. Despite the fact that both \glspl*{gpu} and \glspl*{fpga} are suitable for parallel applications, \glspl*{gpu} have a well-defined architecture, whereas \glspl*{fpga} are flexible devices that allow users to describe new hardware architectures, such as brain-like and neuromorphic architectures \cite{duarte2015}. Nevertheless, \glspl*{gpu} have also proven to be suitable to implement \gls*{snn} architectures \cite{knight2018}. In this work, a \gls*{fpga} device was used to accelerate a customised 3-layer \gls*{snn}.

The \acrfull*{neurohsmd} is the reconfigurable hardware implementation of the \gls*{hsmd} algorithm \cite{Machado2021}. A high-end Intel Stratix 10 \gls*{fpga}(see section~\ref{Ch3:hardware_platform} for further details) was selected for accelerating the \gls*{hsmd}'s \gls*{snn}. \acrfull*{opencl} was used for describing the \gls*{snn} because it provides a higher level of abstraction when compared with \glspl*{hdl}, increased productivity and compatibility with other \gls*{opencl} platforms (such as other \gls*{fpga} devices, \glspl*{gpu} and \glspl*{cpu}). 

The paper is structured as follows: the background research is presented in Section \ref{Ch2:background research}, the hardware platform details are discussed in Section~\ref{Ch3:hardware_platform}; the \gls*{neurohsmd} architecture is presented in Section~\ref{Ch4:NeuroHSMD}; the results are presented in Section~\ref{Ch5:results} and the discussion of the \gls*{neurohsmd} results and future work are in Section~\ref{Ch6:conclusion}.

\section{Background research}\label{Ch2:background research}
\begin{wrapfigure} {l}{0.5\textwidth}
	\begin{center}
	\includegraphics[width=0.5\textwidth]{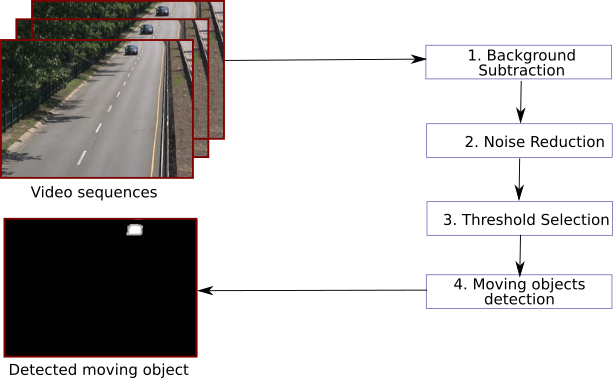}
	\caption{\acrlong{omd} steps.} \label{fig:omd-steps}
	\end{center}
\end{wrapfigure}

The detection of moving objects from video frame sequences is a trivial visual task performed by vertebrate retinal \gls*{gc} \cite{Gollisch2010,Kolb2003} and yet a challenge in the \gls*{cv} research field. \gls*{omd} is one of the most researched fields in computer vision and has been studied for more than 30 years \cite{Chapel2020}. \gls*{omd} in videos captured from static and/or moving cameras is essential for a wide range of computer vision applications such as video surveillance, object collision avoidance, \gls*{adas}, etc \cite{KumarBheemavarapu2021,Garcia-Garcia2020,Chapel2020}. Although the initial \gls*{omd} models were designed for static cameras, the advances in sensor technology and the accessibility to portable devices fitted with cameras is triggering more challenging scenes where both cameras and objects can move at the same time \cite{Chapel2020}. \gls*{omd} includes the following tasks: 1) \acrlong{bs}, 2) noise reduction, 3) threshold selection and 4) moving objects detection (see Figure~\ref{fig:omd-steps}).

Several challenges have been identified in various works \cite{bouwmans2010,bouwmans2014,bouwmans2014book,Bouwmans2019,Garcia-Garcia2020,Chapel2020} and can be summarised as follows : 1) \textbf{Bootstrapping}: the sequence of images includes objects in both the background and foreground; 2) \textbf{Camouflage}: the objects in the foreground are either obstructed by background objects or are composed of similar colours; 3) \textbf{Dynamic background}: the objects in the background include parasitic movements such as surface water movement, branches and leafs shaking in trees, flags on windy days, etc; 4) \textbf{Camera aperture}: undesired blurred background and foreground as a consequence of the incorrect opening in a lens through which light passes to enter the camera; 5) \textbf{Variation of illumination}: instant variations of illumination will increase the number of false-positive detections (i.e. pixels that should belong to the background are classified as foreground); 6) \textbf{Low frame rate}: the temporal distance between image frames prevents instant updates of the background and illumination changes, which reduces the accuracy and increases the number of false positives; 7) \textbf{Motion blur}: caused by rapid camera movements or jittering, which blurs the image;
8) \textbf{Parallax}: the apparent displacement of an object as a consequence of the camera movement. The Parallax will have implications on the background modelling and its compensation; 9) \textbf{Moving camera}: moving cameras introduce complexity because the static objects seem to be moving, and objects moving at a similar speed in the same direction of the camera will seem to be static; 10) \textbf{Background objects movement}: although static objects can be added to and removed from the background, such objects should still be considered static; 11) \textbf{Night videos}: night videos have dim light, lower contrast and reduced colour information; 12) \textbf{Noisy images}: low-quality sensors, dust exposure, dirty lens, bright lights and low resolution are examples of factors that cause noisy images; 12) \textbf{Shadows}: shadows created by objects when exposed to light sources (e.g. sun rays and artificial illumination) should not be part of the foreground models; 13) \textbf{Stationary foreground objects}: a foreground that has stopped moving for a short period should not become part of the background model; 14) \textbf{Challenging weather}: weather conditions (such as fog, rainstorms, strong winds, intense sun rays) have a major impact on the image quality and reduce the quality of the image drastically.

\subsection{Background Subtraction}
Several surveys about \gls*{bs} have been published in the literature focused on static or semi-static (i.e. cameras fixed in a given position exhibiting pan-tilt-zoom movements) scenes. McIvor \cite{mcivor2000} published, in 2000, one of the first \gls*{omd} surveys where nine \gls*{bs} methods which were only described in detail but not compared. Piccardi \cite{piccardi2004} presented, in 2004, a comparative study between seven algorithms considering speed, memory resources utilisation and accuracy (see Table~\ref{tab:bs_analysis}). Piccardi's study \cite{piccardi2004} aims to facilitate the \gls*{bs} selection based on speed, memory requirements and accuracy requirements. 

\begin{table}[ht] \caption{\gls*{bs} methods and their performance analysis} \label{tab:bs_analysis}
\resizebox{14.5cm}{!}{\begin{tabular}{|l|c|c|c|}
\hline\hline
Method & Speed & Memory & Accuracy \\ \hline
Running Gaussian average \cite{wren1997,koller1994}& high & low & acceptable \\ \hline
Temporal median filter \cite{lo2001,cucchiara2003} & high & low & acceptable \\ \hline
Mixture of Gaussians \cite{Stauffer1999} & low & high & very good \\ \hline
Kernel density estimation \cite{han2004}  & low & high & very good \\ \hline
Sequential kernel density approximation \cite{elgammal2000} & low & acceptable & good \\ \hline
Cooccurence of image variations \cite{seki2003} & acceptable & acceptable & good \\ \hline
Eigenbackgrounds \cite{oliver2000} & acceptable & acceptable & good \\ \hline
\end{tabular}}
\end{table}
Cheung et al. \cite{cheung2005} proposed a method for validating foreground regions (blobs) using a slow-adapting Kalman filter and compared the proposed method against six other methods using the recall and precision metrics. Elhabian et al. \cite{elhabian2008} covered several background removal algorithms and identified that all the \gls*{bs} algorithms follow four significant steps, namely, pre-processing, background modelling, foreground extraction, and validation. Although the review was very comprehensive, the focus was on recursive and non-recursive approaches, which are suitable for background maintenance but less suitable for background modelling. Cristiani et al. \cite{cristani2010} reviewed \gls*{bs} methods that can be applied to data captured from different sensor channels (including audio). Elgammal \cite{elgammal2014} reviewed more than 100 papers about object motion detection for static and moving cameras, highlighting the challenges and suggesting which method to use in each case. Bouwmans et al., Garcia et al. and Chapel et al. published comprehensive surveys \cite{bouwmans2010,bouwmans2014,bouwmans2014book,Bouwmans2019,Garcia-Garcia2020,Chapel2020} focusing on traditional, recent, and prospective object motion detection methods.

\begin{wrapfigure} {l}{0.5\textwidth}
	\begin{center}
	\includegraphics[width=0.5\textwidth]{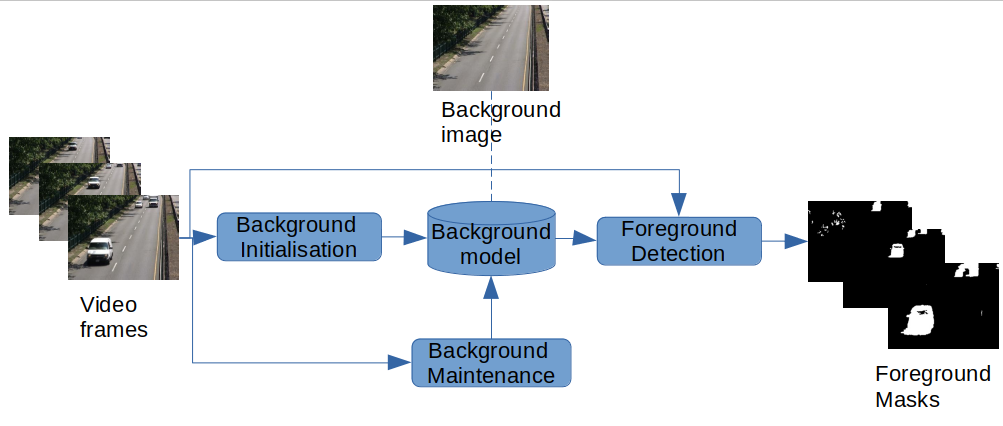}
	\caption{\acrlong{bs} steps.} \label{fig:bs-steps}
	\end{center}
\end{wrapfigure}

Consecutive frame difference, background modelling and optical flow are the main categories for \gls*{bs}. Consecutive frame difference methods are the simplest to implement and require less computational resources, but are also the most sensitive to the challenges listed above \cite{Chapel2020,Garcia-Garcia2020}. In contrast, optical flow methods are the most robust but require more computational resources and, consequently, are not suitable for real-time applications \cite{Chapel2020,Garcia-Garcia2020}. Therefore, background modelling methods are commonly used methods for extracting the foreground from the background in real-time applications \cite{Chapel2020,Garcia-Garcia2020}. \gls*{bs} generic steps are detailed in Figure~\ref{fig:bs-steps}.

Stauffer \& Grimson \cite{Stauffer1999}, and KaewTraKulPong \& Bowden \cite{KaewTraKulPong2002} suggested modelling each pixel as a \gls*{mog} where the Gaussian distributions of the adaptive mixture model are analysed for determining which ones are likely to belong to the background process. All the pixel values that do not fit in the background distributions are considered foreground \cite{Stauffer1999}. Zivkovic \cite{Zivkovic2004} proposes an efficient adaptive algorithm using the \gls*{mog2} for enhancing the \gls*{mog} algorithm. \gls*{mog2} selects automatically the number of components per pixel, which results in complete adaptation to the observed scene. Zivkovic \& Heijden \cite{Zivkovic2006} identified recursive equations for updating the parameters of the \gls*{mog} and proposed \gls*{knn} for the automatic selection of the pixel components. The Gaussian mixture based algorithms (\gls*{mog}, \gls*{mog2} and \gls*{knn}) show good robustness when exposed to noise and losses due to image compression but lack sensitivity to intermittent object motion, moving background objects and abrupt illumination changes.

In 2016, Sagi Zeevi \cite{CNT-2016} proposed the CNT algorithm, which performed better on the \acrfull*{cdnet2014} dataset \cite{Wang2014} and targets embedded platforms (e.g. Raspberry PI\footnote{Available online, \protect\url{https://www.raspberrypi.com/products/raspberry-pi-4-model-b/}, last accessed 28/03/2022}). The CNT uses minimum pixel stability for a specified period for modelling the background; this can vary from 170 ms, default for swift movements, up to hundreds of seconds, where 60s is the default value \cite{Zeevi2016}. Godbehere \textit{et al.} \cite{Godbehere2014} suggested a single-camera statistical segmentation and tracking algorithm named the GMG by combining per-pixel Bayesian segmentation, a bank of Kalman filters, and Gale-Shapley matching for the approximation of the solution to the multi-target problem. The proposed GMG algorithm is limited when processing video streams susceptible to camouflage, losses due to image compression, and noise.

Guo \textit{et al.} \cite{Guo2016} reported an adaptive \acrlong{bs} model enhanced by a local \gls*{lsbp} for addressing illumination changes. The proposed \gls*{lsbp} algorithm enhances the robustness of the motion detection to illumination changes, shadows, and noise. However, the \gls*{lsbp} is less effective when processing video streams susceptible to camouflage, losses due to image compression, or external noise. More recently, in 2017, \gls*{opencv} released an improved version of the \gls*{lsbp} algorithm, also known as \gls*{gsoc} \cite{GSOC2020,samsonovcode2017}, developed during the Google Summer of Code event \cite{Google2017}, which enhances the \gls*{lsbp} algorithm by using colour descriptors and stabilisation heuristics for motion compensation \cite{Samsonov2017,samsonovcode2017}. The \gls*{gsoc} algorithm demonstrates better performance on the \acrfull*{cdnet2012} \cite{Goyette2012}, and \gls*{cdnet2014} \cite{Wang2014} datasets \cite{Samsonov2017,OpenCV2021} when compared to other algorithms available on the \gls*{opencv} library.\\

More recently, Braham et al. \cite{braham2017} proposed a \gls*{sbs} that uses object-level semantics to meet a range of problematic background subtraction conditions. The proposed \gls*{sbs} reduces false positive detections by integrating the output information of a semantic segmentation method, expressed as a probability for each pixel, with the output of existing \gls*{bs} methods. Inspired by Braham's work \cite{braham2017}, Zeng et al. \cite{zeng2019} proposed a \gls*{rtss} for performing \gls*{bs}. The \gls*{rtss} consists of two components: a \gls*{bs} segmenter B and a semantic segmenter S that work in parallel for foreground segmentation. The \gls*{rtss} achieves state-of-the-art performance among most unsupervised background subtraction methods while functioning in real-time as compared to other \gls*{bs} methods \cite{zeng2019}. Liang et al. \cite{liang2018} proposed a deep background subtraction method using a directed learning strategy that learns a specific \gls*{cnn} model for each video without manually labelling. Zeng et al. \cite{zeng2018} proposed a \gls*{mfcn} architecture for background subtraction that takes advantage of diverse layer features. The deep features learned from \gls*{mfcn} improves foreground detection, and the complexity of the background subtraction process can be easily handled during the subtraction operation itself.

\gls*{bs} can be done using methods based on signal processing, \gls*{ml}, \gls*{dnn} or mathematical models. Although signal processing, \gls*{ml} and \gls*{dnn} tend to exhibit better accuracy than mathematical models, these algorithm types are also computationally intensive, introducing undesirable latencies. Nevertheless, mathematical models exhibit lower accuracy but require fewer computational resources and, therefore, are suitable for real-time applications. Moreover, the methods proposed by Braham et al. \cite{braham2017} and Zeng et al. \cite{zeng2019b} demonstrated that existing \gls*{bs} algorithms can be improved when combined with semantic segmentation models. Therefore, in this work, we propose an approach to accelerate the \gls*{hsmd}'s \gls*{snn} which was identified to be the bottleneck of the \gls*{hsmd} algorithm \cite{Machado2021}.

\subsection{Spiking Neural Networks}
\glspl*{fpga} have been used, for many decades, accelerating applications, including edge/cloud computing. \glspl*{fpga} are flexible devices because of their flexible architecture enables developers to describe customised architectures. Such flexibility comes with a downside because \glspl*{fpga} are also known by their associated complexity. There are two main \gls*{fpga} manufacturers, namely, Intel\footnote{Available online, \protect\url{https://www.intel.co.uk/content/www/uk/en/products/programmable/fpga.html}, last accessed: 04/03/2021} and Xilinx\footnote{Available online, \protect\url{https://www.xilinx.com/}, last accessed: 04/03/2021}. \\
Mishra et al. \cite{Misra2010} identified in their survey that many \gls*{snn} usually have about $10^4 \sim 10^8$ neurons and $10^{10} \sim 10^{14}$ synapses and that high-performance neural hardware is essential for practical application. Li et al. \cite{Li2010} proposed the implementation of visual cortex neurons on \glspl*{fpga}. The implemented visual cortex neurons exhibited the same dynamics as those recorded from real neurons using multi-electrodes arrays. Li et al.\cite{Li2012} implemented 256 fully connected neurons, and its performance was assessed by storing four patterns and applying similar patterns containing errors. The implemented system was capable of operating using a 100 MHz clock, which enables the acceleration of the system 40 times above the real-time operation \cite{Li2012}. Cassidy et al. \cite{Cassidy2013} proposed the use of \glspl*{fpga} to accommodate spiking neurons and unsupervised \gls*{stdp} learning structures. In this work, Cassidy et al. \cite{Cassidy2013} demonstrated that digital neuron abstraction is preferable to more realistic analogue neurons; they also emulated the massive parallelism connectivity and high neuron density as observed in nature; the neuron states were also multiplexed to take advantage of clock frequencies and dense \glspl*{sram}. 
Chen et al. \cite{chen2017a} described a \gls*{cpg} composed of two reciprocally inhibitory neurons. To reduce the \gls*{fpga} resources usages, Chen et al. \cite{chen2017a} has optimised the \gls*{cpg} to avoid using multipliers (\glspl*{fpga} have a low quantity of multiplier blocks), and the non-linear parts of the Komendantov-Kononenko neuron model \cite{komendantov1996} were removed. Cheung et al. \cite{Cheung2016} proposed the NeuroFlow, a scalable \gls*{snn} simulator suitable to be implemented on \gls*{fpga} clusters. It was possible to simulate about 600,000 neurons and to get a real-time performance for up to 400,000 neurons simulated using NeuroFlow on 6 \glspl*{fpga} \cite{Cheung2016}. Podobas and Matsuoka \cite{Podobas2017} proposed the use of \gls*{opencl}, an \gls*{hls} tool, to increase productivity by facilitating the \gls*{snn} design (provide a higher level of hardware abstraction) on \glspl*{fpga}. Two different neuron models, their axons and synapses, were designed using \gls*{opencl} and the authors claim a speed performance of up to 2.25 GSpikes/second. Sakellariou et al. \cite{Sakellariou2021} suggested a spiking accelerator base on \glspl*{fpga} to enable users to develop \glspl*{snn} targeting \gls*{ml} applications and promise an acceleration of up to 800 times for inference and up to 500 times for training compared to Software \gls*{snn} simulations.

\begin{figure}
\centering
\includegraphics[width=1.0\textwidth]{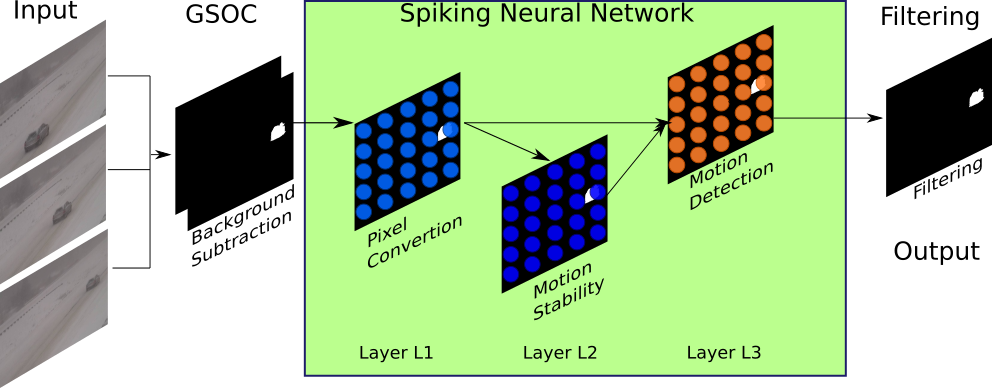}\caption{\gls*{hsmd} architecture implemented by Machado et al. \protect\cite{Machado2021}. The \gls*{gsoc} algorithm performs the background subtraction, the customised \gls*{snn} performs the \gls*{omd} and the filter removes noise. The customised \gls*{snn} is composed of 3 layers for 1) converting pixel values conversion into currents, 2) motion stability and 3) perform motion detection.}\label{fig:hsmd}
\end{figure}

Machado et al. \cite{Machado2021} proposed the \gls*{hsmd} model (see Figure~\ref{fig:hsmd}) inspired by the object motion functionality exhibited by vertebrate retinas, in which \gls*{oms-gc} determine the difference between a local patch's motion trajectory and the background \cite{Gollisch2010}. The \gls*{hsmd} uses a 3-layer \gls*{snn} to enhance the \gls*{gsoc} \gls*{bs} algorithm \cite{Samsonov2017,samsonovcode2017,Machado2021}.

The works reviewed in this section demonstrated that \glspl*{fpga} offer flexibility, high efficiency, low-power, and high degree of parallelism, making \glspl*{fpga} is the most suitable devices for implementing brain-like circuits. \glspl*{fpga} enable the design of complex biological plausible neuron models and massively parallel \gls*{snn} composed of thousands of \gls*{lif} neurons capable of generating complex biological like patterns. Although \gls*{fpga} devices are normally programmed using complex \gls*{hdl} tools, \gls*{hls} tools such as \gls*{opencl} can be used to increase the productivity of the \glspl*{fpga} design process by providing hardware abstraction which reduces the implementation complexity. In this paper, we utilise \gls*{opencl} to design the complex \gls*{snn} architecture of the \gls*{hsmd}. The \gls*{neurohsmd} reported in this paper, improves the speed of the \gls*{hsmd} \cite{Machado2021} without degradation of the background subtraction accuracy using an high-end \gls*{fpga} device.\\

\section{Hardware Platform} \label{Ch3:hardware_platform}
The \gls*{hsmd} algorithm \cite{Machado2021} has proven to be very sensitive to object motion events triggered by objects, as a direct consequence of using an \gls*{snn} to emulate the basic functionality observed in \gls*{oms-gc}. The \gls*{snn} utilised is composed of 4 layers of neurons interconnected on a 1:1 synaptic connectivity. 

\subsection{Field Programmable Gate Array}
The target \gls*{fpga} board is fitted with a state-of-the-art Stratix 10 SoC \gls*{fpga} device\footnote{Available online, \protect\url{https://www.intel.co.uk/content/www/uk/en/products/programmable/soc/stratix-10.html}, last accessed: 07/04/2021}. The Intel Stratix family is composed of \glspl*{lab} made of 10 basic building blocks called \glspl*{alm}. Each \gls*{alm} is composed of fractionable Look-Up-Tables, also known as \gls*{alut}, two-bit full adder and four registers. \glspl*{lab} can be freely reconfigured to implement logic and arithmetic functions. Furthermore, up to a quarter of the \glspl*{lab} can be used as \glspl*{mlab}. Each \gls*{lab} contains dedicated logic elements used to driving control signals to \glspl*{alm}. Each \gls*{mlab} supports up to 640 bits of simple dual-port \gls*{ram}. It is possible to configure each \gls*{alm} in an \gls*{mlab} as $32 \times 2$ memory blocks equivalent to $32 \times 2 \times 10$ simple dual-port \gls*{ram} blocks. Dual-port \glspl*{ram} are low-latency memory devices that only takes a clock cycle to perform a read/write operation (for example, \gls*{sdram} in \glspl*{cpu} takes thousands of clock cycles to complete read/write operations). Furthermore, the Stratix 10 offer variable-precision \gls*{dsp} blocks that can support fixed-point arithmetic and single-precision floating-point arithmetic.

\subsection{Open Computer Language (\gls*{opencl})} \label{Ch3.4:opencl}
\gls*{opencl} applications are split into two parts, namely, \textbf{host} application(s) and \textbf{device} kernel(s). The \textbf{host} applications are always compiled on the host Operating System and run on a \gls*{cpu}. \textbf{Host} applications are also used to launch the target kernels on the target \textbf{devices}. Kernels are special functions written in \gls*{opencl} C/C++ to perform parallelisable computations on accelerators such as \glspl*{gpu} and \glspl*{fpga} \cite{Intel2019}. For instance, consider two m$\times$n matrices A and B where it is expected to do the operation C=A+B where C is the third matrix of m$\times$n. In this case, the kernel could just perform, in parallel, the addition of matrices A and B elements and store the result in C. Unlike in \glspl*{cpu}, where it would take m$\times$n operations to complete this matrix addition, \glspl*{gpu} and \glspl*{fpga} could parallelise this operation depending on the resources available per device resulting in the acceleration of the application.
Buffer objects within a context are used in \gls*{opencl} to exchange data between the host and device \cite{CodeProjects2011}. The Intel \gls*{fpga} \gls*{sdk} for \gls*{opencl} offline compiler optimises the kernel throughput by adjusting buffer sizes during the kernel compilation process \cite{Intel2021}. \gls*{opencl} provides both mapped and asynchronous buffers, enabling the application to continue to run while additional data is exchanged.
\begin{wrapfigure} {l}{0.5\textwidth}
\centering
\includegraphics[width=0.5\textwidth]{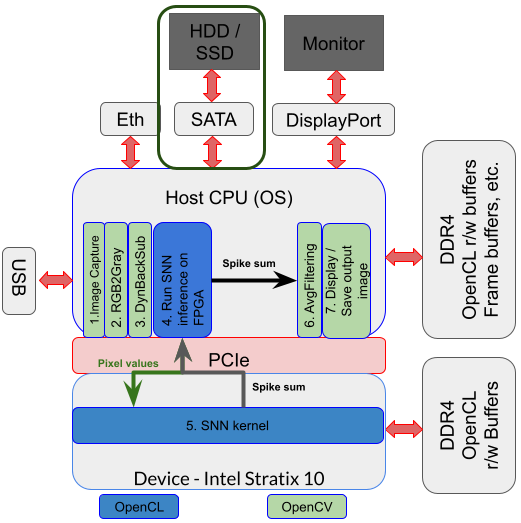}\caption{\gls*{neurohsmd} architecture. The diagram represents the computation stages that run both on the \gls*{cpu} and \gls*{fpga}. Shows that the \gls*{fpga} is connected to the host \gls*{cpu} via the \gls*{pcie} bus. It also shows the dedicated memory of the \gls*{cpu} and \gls*{fpga} device. Includes also how external devices (e.g. cameras, monitor and \gls*{hdd}/\gls*{ssd}) connect to the host \gls*{cpu} via different interfaces (e.g. ethernet (eth), \gls*{sata}, display port and \gls*{usb}). The computations stages implemented in \gls*{opencl} are in blue and \gls*{opencv} in green.} \label{fig:NeuroHSMD_arc}
\end{wrapfigure}
Software Developers have to carefully analyse the code to be optimised and only select the sections that may benefit from the hardware acceleration because the maximum speed is always dictated by the \gls*{pcie} bus speed. Another big challenge for Software Developers is the low debugging capabilities available while the code is being executed on the device.\\

Although it is possible to use \gls*{opencl} to program \gls*{fpga} and \gls*{gpu} devices, \glspl*{gpu} are specialised devices designed for video rendering and graphics processing. At the same time, \glspl*{fpga} are customisable devices that can be freely reconfigurable. Therefore, \glspl*{fpga} offer more flexibility than \glspl*{gpu}, which is desirable for accelerating \glspl*{snn} because they can be modelled using the \glspl*{noc} concept. Each individual spiking neuron can be considered a node that interconnects to one or more nodes (neurons) of the same \gls*{snn}. The flexibility offered by both \glspl*{fpga} and \gls*{opencl} makes the selection of \glspl*{fpga} over \glspl*{gpu} the obvious choice.

The Intel \gls*{fpga} \gls*{sdk} for \gls*{opencl} (IOCL) provides a compiler and powerful tools to build and run \gls*{opencl} applications targeting Intel \gls*{fpga} devices. The IOCL generates two main components: the host application and the \gls*{fpga} programming bitstream(s). The IOCL offline compiler (AOC) first compiles the custom kernel(s) to an image file (*.aocx) that will be used to program the \gls*{fpga}. In contrast, the host-side C/C++ compiler compiles the host application and then links it to the IOCL runtime libraries. Fixed-point data representations, which only retain the necessary data resolution for calculations and can result in hardware savings, are a popular choice among hardware developers. However, the IOCL standard lacks support for fixed-point representation and floating operations must be carried out using IEEE754 single-precision floating-point \cite{Intel2017}. 

The IOCL compiles one or more \gls*{opencl} kernels and creates a hardware configuration file. A successful compilation results in a \textbf{*.aocr}, \textbf{*.aoco}, \textbf{*.aocx} and reports/report.html files. The report.html contains the estimated resource usage and a preliminary assessment of area usage. The intermediary \textbf{*.aoco} and \textbf{*.aocr} are only used in the generation of the \textbf{*.aocx} which is then used to program the \gls*{fpga}.\\
\begin{wrapfigure} {r}{0.5\textwidth}
\centering
\includegraphics[width=0.45\textwidth]{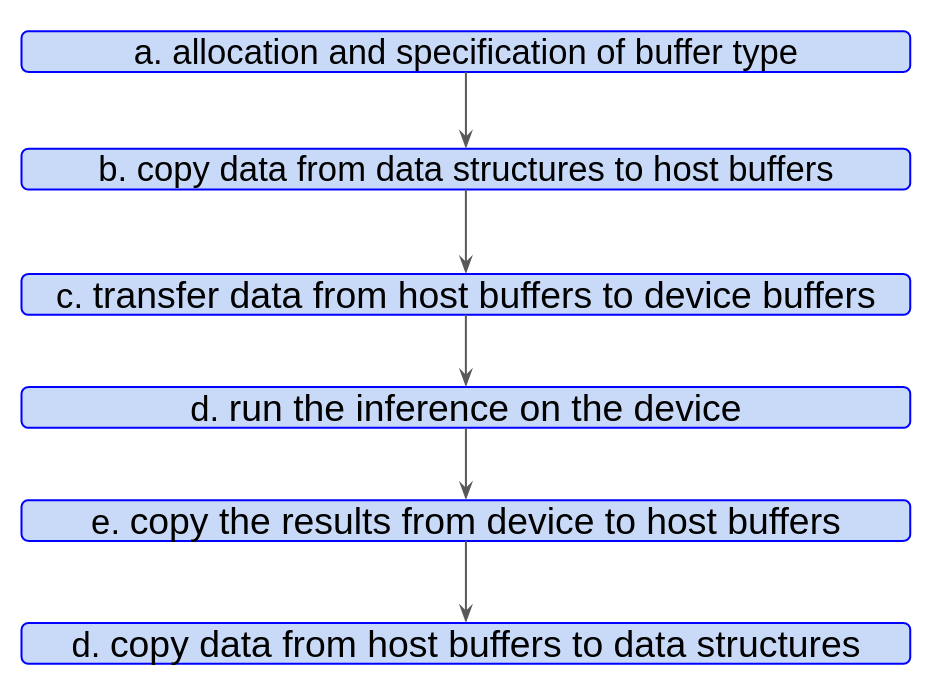}\caption{\gls*{opencl} stages.} \label{fig:opencl_stages}
\end{wrapfigure}
The Terasic DE10-pro development board \cite{Terasic2021} equipped with a state-of-the-art high-end Intel Stratix 10 \gls*{fpga} device was used for implementing the \gls*{neurohsmd} discussed in this section. Terasic states that DE10 pro was designed to fulfil the demands of AI, Data Center, and High-Performance Computing. Furthermore, the DE10-pro development board takes advantage of the latest Intel Stratix 10 to obtain high-speed and low-power (with up to 70\% lower power when compared with the previous generation - i.e. Stratix V). It is equipped with 32GB DDR4 memory module running at over 150 Gbps, up to 15.754 GB/s data transfer via \gls*{pcie} Gen 3 x16 edge between \gls*{fpga} and host workstation, and 4 onboard QSFP28 (100GbE) connectors. The DE10 pro was installed on the host workstation equipped with an Intel(R) Core(TM) i7-4770 \gls*{cpu} @ 3.40GHz and 16 GB of DDR3 using the \gls*{pcie} slot.

\section{NeuroHSMD architecture} \label{Ch4:NeuroHSMD}
\begin{wrapfigure} {l}{0.5\textwidth}
\centering
\includegraphics[width=0.4\textwidth]{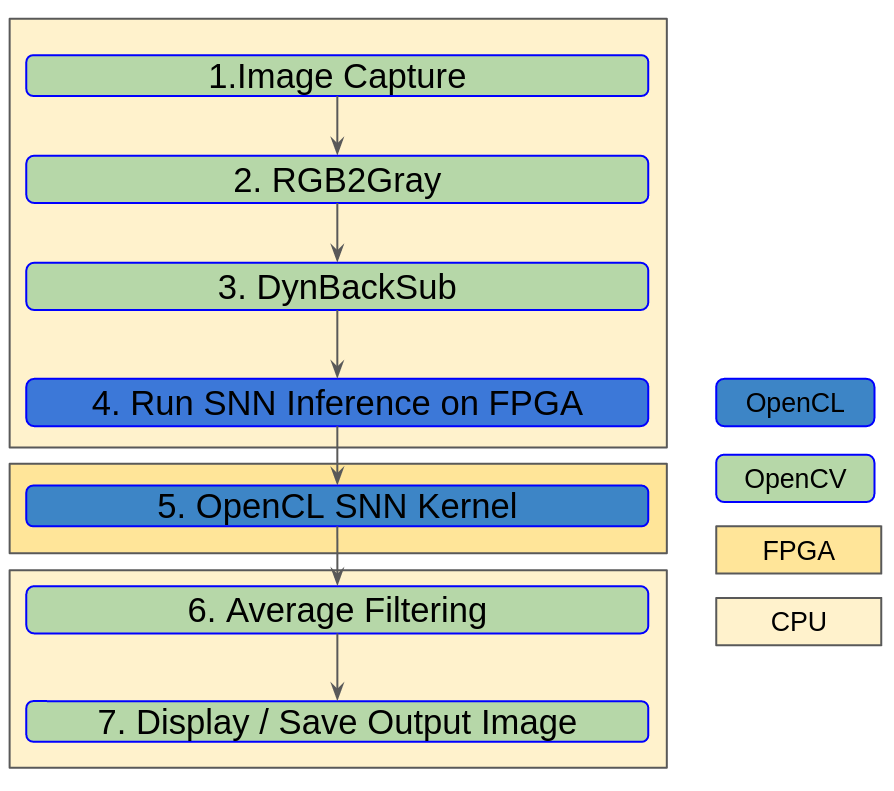}\caption{\gls*{neurohsmd} computation stages.} \label{fig:neurohsmd_stages}
\end{wrapfigure}
The \gls*{neurohsmd} architecture is described in this section. In \gls*{opencl}, the data exchanged between the host application and the \gls*{fpga} kernels flow is as follows: a) allocation and specification of buffer type (i.e. read or write) on the host and device; b) copy data from the application data structures to host buffers; c) transfer data from host buffers to device buffers; d) run the inference on the device; e) copy the results from device to host buffers; f) copy data from the host buffers to application data structures (see Figure~\ref{fig:opencl_stages}).

The \gls*{neurohsmd} algorithm performs the following stages of computation: 1) image capture, 2) conversion from colour to grey, and 3) dynamic background subtraction using the \gls*{opencl}'s \gls*{gsoc} algorithm and copy resulting pixel values are buffered and transferred to the \gls*{fpga} device; 4) run the inference on the \gls*{fpga} and wait for the spike results; 5) run the \gls*{snn} kernel, 6) Filtering using an average filter, and
 7) Display and save the output image.
The \gls*{neurohsmd} computation stages are summarised in Figure \ref{fig:neurohsmd_stages} and the \gls*{neurohsmd} architecture is depicted in Figure~\ref{fig:NeuroHSMD_arc}.\\

Furthermore, the NeuronHSMD \gls*{opencl} is composed of the \gls*{nha} and the \gls*{ndk}. Details about the \gls*{nha} and \gls*{ndk} are given in sections~\ref{Ch3.6.1:host} and \ref{Ch3.6.2:kernels}. 

\subsection{Spiking Neuron Model} \label{ch4.1:neurons}
This study employed the \gls*{lif} spiking neuron model due to its simplicity, computational efficiency, and applicability for near-real-time picture processing. When compared to real biological neurons, the \gls*{lif} spiking neuron model has comparable but less complex dynamics \cite{Gerstner2002}. There are additional sophisticated spiking neuron models, such as Hodgkin-Huxley, however these need substantial computational resources and have a larger effect on computational performance (e.g. Izhikevich \cite{Izhikevich2004}). The dynamics of the \gls*{lif} neuron follow equation \ref{eq:1} \cite{Gerstner2002}.
\begin{eqnarray}
\label{eq:1}
I(t)=\frac{Vm(t)-Vm_{rest}}{R}+C\frac{\partial Vm}{\partial t} \\
\nonumber
\end{eqnarray}
\begin{eqnarray}
\label{eq:2}
\tau_{m}\frac{\partial Vm}{\partial t}=-[Vm(t)-Vm_{rest}]+RI(t)\\
\nonumber
\end{eqnarray}

Where $C$ is the membrane capacitance, $R$ is the membrane resistance, $I(t)$ is the current in a given time t, $Vm(t)$ is the membrane potential in a given time t, $\tau_{m}$ is constant given by the resister $R$ times the capacitor $C$ and the $Vm_{rest}$ is the reset potential.

The action potential $t(f)$ that is emitted by a given neuron is known as the firing time. The firing time $t^{(f)}$ is given by equation \ref{eq:3} \cite{Gerstner2002}.

\begin{eqnarray}
\label{eq:3}
\lim_{\delta\rightarrow 0;\delta >0} t^{(f)}: Vm(t^{(f)})= Th \\
\nonumber
\end{eqnarray}
\vspace{-4mm}

The firing time $t^{(f)}$ is generated when the potential is reset to a new value $Vm_r< Th$ where $Th$ is the threshold (see equation \ref{eq:4} \cite{Gerstner2002}).

\begin{eqnarray}
\label{eq:4}
\lim_{\delta\rightarrow 0;\delta >0} Vm(t^{(f)}+\delta)=Vm_r\\
\nonumber
\end{eqnarray}
\vspace{-4mm}

For $t > t^{(f)}$ the dynamics given by equations \ref{eq:1} and \ref{eq:2} until the next threshold $Th$ crossing occurs. ($\delta$ is the Dirac function). The combination of leaky integration in equations \ref{eq:1} and \ref{eq:2} and reset \ref{eq:3} is given by equation \ref{eq:5} \cite{Gerstner2002}.

\begin{eqnarray}
\label{eq:5}
S_i(t)=\sum _f \delta(t-t_i^{(f)})\\
\nonumber
\end{eqnarray}
\vspace{-4mm}

The Euler method was employed to numerically solve the \glspl*{ode} that model the behaviour of the \gls*{lif} neurons. This method was used to approximate the solutions of the \glspl*{ode} at the simulation's time-steps throughout the simulation. At each time step, the Euler method used the derivative of the current solution to estimate the next solution, effectively advancing the state of the neuron across time.

\subsection{Layers and interconnectivity}\label{ch4.2:layers}

\subsubsection{Input Layer: \gls*{bs} and reduction}\label{input_layer}
\hfill\\
Each $n \times m$ image frame (i.e. camera, video sequence or image sequences) is transformed into greyscale. The \gls*{gsoc} \cite{GSOC2020} provides an adaptive \gls*{bs} utilising colour descriptors and different stabilisation heuristics \cite{Samsonov2017,samsonovcode2017} while computing the frames pixel-by-pixel and exploiting the scalability offered by the \gls*{opencv}  \cite{Samsonov2017}. 

\subsubsection{Layer 2: Pixel intensities to currents encoding}\label{layer_2}
\hfill\\
Pixel intensity values are transformed to proportionate currents and delivered to the spiking neurons in Layer 2 through a 1:1 connection. The Layer 1 neurons were trained to generate spike events proportional to the pixel intensities, and are ruled by the equation~\ref{eq:pixel_current}. 
\begin{equation}
    i_c(x,y)=I(x,y). c
    \label{eq:pixel_current}
\end{equation}

\noindent where $i_c(x,y)$ is the corresponding current for the image light intensity value $I(x,y)$ at coordinates $x$ and $y$, and $c$ is a conversion constant obtained experimentally (in our case, $c$=17.5).

\subsubsection{Layer 3: Motion stability} \label{layer_3}
\hfill\\
\begin{wrapfigure} {l}{0.5\textwidth}
\centering
\includegraphics[width=0.45\textwidth]{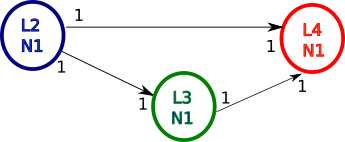}
\caption{\gls*{hsmd} connectivity. In this example, it can be seen that the neuron 1 (N1) of each layer connects to the N1 of the subsequent layer.
\label{fig:connectivity}} 
\end{wrapfigure}
The neurons in Layer 3 are utilised to stabilise motion by creating local buffers and delaying the transmission of spike occurrences.
Neurons in layer 2 connect to Layer 3 neurons and Layer 3 neurons to the Layer 4 neurons for creating local delays. The Layer 3 neurons are used to buffer spike events for one simulation time-step ($\delta t$, in this work, $\delta t=10 ms$) and forwarded to the Layer 4 neurons for the subsequent simulation time-step. The Layer 2 neurons extend 1:1 synaptic connectivities to the Layer 3 neurons and Layer 3 neurons also extend 1:1 synaptic connectivities to the Layer 4 neurons as depicted in Figure~\ref{fig:connectivity}. All the synapses were assigned the value 1370, which was obtained experimentally.

\subsubsection{Layer 4: Motion detection} \label{layer_4}
\hfill\\
The neurons in Layer 4 receive excitatory synaptic connections from the neurons in both Layers 2 and 3 and utilise the spike patterns generated to sense motion.
The generation of spike events by Layer 4 neurons is the result of dynamic changes triggered by successive image frames. Signals received directly from Layer 2 neurons allow for the detection of differences between the current image frame n and the prior image frame n-1. In contrast, Layer 3 neurons are used for comparing image frame n-1 to image frame n-2. Layer 4 spike events are mapped to the matching area of the current image area. The synaptic weight of 1370 for all the synapses was obtained experimentally. The Layer 2 to Layer 4 weights were adjusted to feed-forward all the Layer 2 spike events. While the Layer 3 to Layer 4 synaptic weights were adjusted to generate spike events from the Layer 4 neurons for every pair of two consecutive spike events, the primary objective was to give higher priority to recent spike events (frame [n] - frame [n-1]) and lower priority to older spike events (frame [n-1] - frame [n-2]).

\subsubsection{Layer 5: Filtering} \label{layer_5}
\hfill\\
The spike events matrix of Layer 4 neurons is mapped into a motion matrix $M_d$ of the same dimensions as the current image frame (i.e., $ntimes m$). The events in the $M_d$ matrix are filtered via an averaging filter ruled by the following equations \ref{eq:filter} and \ref{eq:conv}:

\begin{equation}
H(u,v)=\frac{1}{u.v}\begin{pmatrix}
\begin{bmatrix}
w_{0,\,0} & ... & w_{0,\,u} \\ 
 ... & ...  & ... \\ 
w_{v,\,0} & ... & w_{v,\,u}
\end{bmatrix}
\end{pmatrix}
\label{eq:filter}
\end{equation}

\begin{equation}
Y_d(x,y)=M_d(x,y)*H(u,v)
\label{eq:conv}
\end{equation}

\noindent where $Y_d(x,y)$ is the filtered motion detection matrix, $H(u,v)$ is the averaging filter, $u$ and $v$ are the convolution window length and height respectively, $*$ is the convolution operator, $w$ is the filter window.

\subsection{Neuronal parameters} \label{neuronal_parameters} 
\hfill\\
The \gls*{snn} was parameterised using the recommended settings in the references \cite{Jolivet2004, Brette2005}. Therefore, the simulation was configured with a time step of $ \delta t$=10 ms, $V_m$=-55.0 mV, $E_L$ = -55.0 mV, $C_m$ = 10.0 pF, $R$=1.0 M$\Omega$, $V_{reset}$=-70.0 mV, $V_{min}$=-70.0 ms, $V_{th}$=-70.0 mV, $\tau$=10.0 ms, $t_{ref}$=2 ms, $w_{syn}$ = 1370 (neurons L3 and L4) and $w_{p2i}$=8.0 (L2 neurons only).

\subsection{Host application} \label{Ch3.6.1:host}  \hfill \break
The \gls*{nha} is used to interface two \glspl*{ndk}, one with implements the \gls*{hsmd} and a second version that includes a speed optimisation (see next section for further details).
Algorithm~\ref{alg:host} summarises each of the computation stages that occur in the \gls*{nha}. 
The communication between the \gls*{nha} and the \gls*{ndk} is limited by the \gls*{pcie} bus speed (~16 GB/s \footnote{Available online, \protect\url{https://www.trentonsystems.com/blog/pcie-gen4-vs-gen3-slots-speeds}, last accessed: 21/06/2021}). In the \gls*{hsmd}, the limitations are only dictated by the \gls*{cpu} speed and the DDR4 memory speed (~34.1 GB/s\footnote{Available online, \protect\url{https://www.crucial.com/support/articles-faq-memory/understanding-cpu-limitations-with-memory}, last accessed: 21/06/2021}) which is 2 times faster than the \gls*{pcie} bus speed. 
\newpage 

\subsection{Device kernels}\label{Ch3.6.2:kernels}  
\hfill \break
\begin{wrapfigure} {l}{0.5\textwidth}
\begin{minipage}[l]{0.46\textwidth}
\begin{algorithm}[H]
\SetAlgoLined
\KwIn{\\
img: image frame;}
\KwOut{\\
post\_proc\_img:\ post\ processed\ image;\\
stats:\ computation\ statistics;}
\KwAlg{
\begin{algorithmic} [1]
 \STATE initialise\_opencl()\;

 \FOR{$iterator\gets list\_folders.begin()$ \TO list\_folders.end()}
 \STATE $(x,y) \gets get\_image\_size()$; 
 \STATE $num\_layers \gets 3$\;
 \STATE $tot\_neurons \gets x.y.num\_layers$\;
 \STATE reset\_opencl\_buffers(tot\_neurons )\;
 \STATE $gsoc \gets initialise\_gsoc\_back\_subtraction$\;
 \FOR{$iterator2\gets files\_list.begin()$ \TO files\_list.end()}
 \STATE $img \gets read\_image(iterator2);$\;
 \STATE $<pixel\_values> \gets gsoc.compute(img)$\;
 \STATE \textbf{NeuroHSMDv1($<pixel\_values>$) $\rightarrow <spike\_sum>$}\; \COMMENT{\protect\textbf{Alg.~\ref{alg:NeuroHSMD-v1}}}
 \STATE \textbf{OR}
 \STATE \textbf{NeuroHSMDv2($<pixel\_values>$) $\rightarrow <spike\_sum>$}\; \COMMENT{\protect\textbf{Alg.~\ref{alg:NeuroHSMD-v2}}}
 \STATE $post\_proc\_img \gets get\_spikes\_sum\_l3(<spike\_sum>)$ \;
 \STATE $save(post\_proc\_img)$ \;
 \STATE $stats \gets compute\_stats(time)$ \;
 \ENDFOR
 \STATE $save(stats)$ \;
 \ENDFOR
 \end{algorithmic}}
\caption{\gls*{neurohsmd} host application.}\label{alg:host}
\end{algorithm} 
\end{minipage}
\end{wrapfigure}
The \gls*{nha} performs the background subtraction using the \gls*{gsoc} algorithm, performs inference of the \gls*{snn} kernel (described on the \gls*{fpga}), and uses the inference results to compute the \gls*{bs}. Furthermore, the \gls*{nha} can process both live images captured from camera devices or extracted from videos stored on \gls*{usb} or \gls*{sata} devices.\\

\begin{wrapfigure} {l}{0.5\textwidth}
\begin{minipage}[l]{0.46\textwidth}
\centering
\includegraphics[width=1.\textwidth]{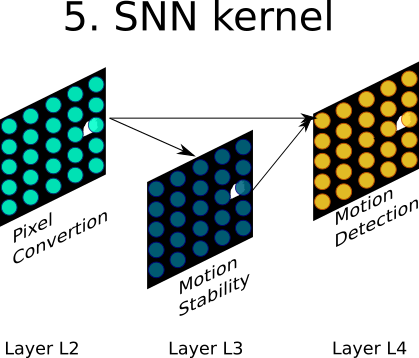}\captionof{figure}{\glspl*{ndk} implementation. The \gls*{hsmd}'s \gls*{snn} includes: Layer L2 - conversion of pixel values to currents, Layer L3 - motion stability and Layer L4 - motion detection} \label{fig:NeuroHSMD_snn}
\end{minipage}
\end{wrapfigure}

The \glspl*{ndk} section covers the two kernels (i.e. NeuroHSMDv1 and NeuroHSMDv2) that have been implemented. The Intel Quartus via the Intel \gls*{fpga} \gls*{sdk} for \gls*{opencl} optimised the device kernels to use of the variable-precision \gls*{dsp} blocks, offering IEEE754 single-precision floating point resolution required for performing the target operations. Both kernels use the IEEE754 single-precision floating point representation, and therefore all computations were performed using single-precision arithmetic. The NeuroHSMDv1 is the equivalent implementation to the HSMDv1 where the spike sum per neuron is computed for all the neurons, while NeuroHSMDv2 is the equivalent to HSMDv2 where the spike sum per neuron is only computed for neurons that have pixel intensity values greater than 0.0. Both, NeuroHSMDv1 and NeuroHSMDv2 \glspl*{ndk} implement the \gls*{hsmd}'s \gls*{snn} (see Figure~\ref{fig:NeuroHSMD_snn}). Furthermore, the synaptic weights were stored in vectors and were exchanged via the device and host \gls*{cpu} memory buffers.

The \textit{NeuroHSMDv1} was parallelised by a factor of 16. The parallell factor of 16 was the optimal parallelism coefficient and was obtained experimentally. The \gls*{ndk} version 1 is the equivalent implementation of the \gls*{hsmd} algorithm \cite{Machado2021}  which is referred to as HSMDv1 in this section. The HSMDv2 is an optimised version of the \gls*{hsmd} where spike sums per neuron are only computed for neurons that have pixel intensity values greater than 0.

The Algorithm~\ref{alg:NeuroHSMD-v2}, inferred by the NHP, summarises the computation steps required to compute the spike sum.

Fig. \ref{fig:NeuroHSMD_snn} depicts the seven stages of computation and exhibits the place where each stage occurs (i.e. \gls*{cpu} or \gls*{fpga}).

\begin{minipage}{0.47\textwidth}
\begin{algorithm}[H]
\SetAlgoLined
\setcounter{AlgoLine}{0}
\KwCircuits{16}
\KwConst{\\$R$: membrane\ resistance;\\
$\tau$: membrane time constant;\\
 $dt$: time step;\\
 $p2c$: pixel\ values\ to\ current;\\
 $steps$: number of steps; \\
 $s2c$: spike\ to\ current; \\
 $number\_neurons$: obtained from largest image to be processed}
\KwIn{\\$<pixel\_val>$: pixel values; \\
$num\_neuron\_layer$: number of neurons per layer (one per pixel);}
\KwOut{\\$<spk\_sum>$: spike sum;}
\KwAlg{
\begin{algorithmic} [1]
 \FOR{$neuron\_idx\gets0$ \TO number\_neurons}
 \STATE $I(t) \gets pixel\_val[neuron\_idx].p2c$;\\
 \FOR{$dt1\gets0$ \TO steps}
 \STATE Layer 1:
 \STATE $compute\_V_m\_l1[neuron\_idx](I(t))$;\  \COMMENT{\protect\textbf{Eq.~\ref{eq:1}}} \\
 \STATE $update\_spike\_sum\_l1[neuron\_idx](V_m\_l1)$\ \COMMENT{\protect\textbf{Eq.~\ref{eq:5}}};\\
 \STATE Layer 2:\\
 \STATE $I(t)\_l2 \gets spike\_sum\_l1[[neuron\_idx].s2c$;\\
 \STATE $compute\_V_m\_l2[[neuron\_idx](I(t)\_l2)$;\  \COMMENT{\protect\textbf{Eq.~\ref{eq:1}}}
 \STATE $update\_spike\_sum\_l2[neuron\_idx](V_m\_l2)$;\ \COMMENT{\protect\textbf{Eq.~\ref{eq:5}}};\\
 \STATE Layer 3:\\
 \STATE $I(t)\_l3 \gets spike\_sum\_l2[neuron\_idx].s2c$;\\
 \STATE $compute\_V_m\_l3[neuron\_idx](I(t)\_l2+I(t)\_l3)$;\  \COMMENT{\protect\textbf{Eq.~\ref{eq:1}}}
 \STATE $update\_spk\_sum[neuron\_idx](V_m\_l3)$;\ \COMMENT{\protect\textbf{Eq.~\ref{eq:5}}};\\
 \ENDFOR
 \ENDFOR
 \end{algorithmic}}
\caption{NeuroHSMDv1}\label{alg:NeuroHSMD-v1}
\end{algorithm} 
\end{minipage}
\hfill
\begin{minipage}{0.47\textwidth}
\begin{algorithm}[H]
\SetAlgoLined
\setcounter{AlgoLine}{0}
\KwCircuits{16}
\KwConst{\\$R$: membrane\ resistance;\\
$\tau$: membrane time constant;\\
 $dt$: time step;\\
 $p2c$: pixel\ values\ to\ current;\\
 $steps$: number of steps; \\
 $s2c$: spike\ to\ current; \\
$number\_neurons$: obtained from largest image to be processed}
\KwIn{\\$<pixel\_val>$: pixel values; \\
$num\_neuron\_layer$: number of neurons per layer (one per pixel);}
\KwOut{\\$<spk\_sum>$: spike sum;}
\KwAlg{
\begin{algorithmic} [1]
 \FOR{$neuron\_idx\gets0$ \TO number\_neurons}
 \IF{\HiLi$pixel\_val[neuron\_idx]>0.0$}
 \STATE $I(t) \gets pixel\_val[neuron\_idx].p2c$;\\
 \FOR{$dt1\gets0$ \TO steps}
 \STATE Layer 1:
 \STATE $compute\_V_m\_l1[neuron\_idx](I(t))$;\ \COMMENT{\protect\textbf{Eq.~\ref{eq:1}}}\\
 \STATE $update\_spike\_sum\_l1[neuron\_idx](V_m\_l1)\ \COMMENT{\protect\textbf{Eq.~\ref{eq:5}}}$;\\
 \STATE Layer 2:\\
 \STATE $I(t)\_l2 \gets spike\_sum\_l1[[neuron\_idx].s2c$;\\
 \STATE $compute\_V_m\_l2[[neuron\_idx](I(t)\_l2)$;\ \COMMENT{\protect\textbf{Eq.~\ref{eq:1}}}\\
 \STATE $update\_spike\_sum\_l2[neuron\_idx](V_m\_l2)$;\ \COMMENT{\protect\textbf{Eq.~\ref{eq:5}}};\\
 \STATE Layer 3:\\
 \STATE $I(t)\_l3 \gets spike\_sum\_l2[neuron\_idx].s2c$;\\
 \STATE $compute\_V_m\_l3[neuron\_idx](I(t)\_l2+I(t)\_l3)$;\ \COMMENT{\protect\textbf{Eq.~\ref{eq:1}}}\\
 \STATE $update\_spk\_sum[neuron\_idx](V_m\_l3)$;\ \COMMENT{\protect\textbf{Eq.~\ref{eq:5}}};\\
 \ENDFOR
 \ENDIF
 \ENDFOR
 \end{algorithmic}}
\caption{NeuroHSMDv2}\label{alg:NeuroHSMD-v2}
\end{algorithm} 
\end{minipage}
\hfill \break

\newpage
Similar to NeuroHSMDv1, the \textit{NeuroHSMDv2} was parallelised by a factor of 16. The parallel factor of 16 was the optimal parallelism coefficient and was obtained experimentally. The \gls*{ndk} version 1 contains an optimisation where the spike sum for a given neuron of layer 1 is only computed if the pixel intensity value is greater than 0.0. This optimisation was also applied to the original \gls*{hsmd} Algorithm \cite{Machado2021}. The optimised version of the \gls*{hsmd} is called HSMDv2 in this section.

\subsection{Datasets and Benchmarking} \label{Ch3.7:datasets_comparison}
The NeuroHSMDv1, NeuroHSMDv2, HSMDv1 and HSMDv2 were tested against the \gls*{cdnet2012} \cite{Goyette2012} and \gls*{cdnet2014} \cite{Wang2014}.

The average performance obtained for each category using each \gls*{bs} method and the \gls*{hsmd} and \gls*{neurohsmd} algorithms are characterised via the eight metrics, as shown below. The four base qualitative metrics are: True Positive (TP), True Negative (TN), False Positive (FP) and False Negative (FN) \cite{Goyette2012,Wang2014}.

\gls*{re}: $Re=\frac{TP}{TP+FN}$

$Re$: ranked by \textbf{descending order};

\gls*{sp}: $Sp=\frac{TN}{TN+FP}$;

$Sp$ ranked by \textbf{descending order};

\gls*{fpr}: $FPR=\frac{FP}{FP+TN}$;

$FPR$ ranked by \textbf{ascending order};

\gls*{fnr}: $FNR=\frac{FN}{FN+TP}$;

$FNR$ ranked by \textbf{ascending order};

\gls*{wcr}:\\ $WCR=\frac{FN+FP}{TP+FN+FP+TN}$;

$WCR$ ranked by \textbf{ascending order};

\gls*{ccr}:\\ $CCR=\frac{TP+TN}{TP+FN+FP+TN}$;

$CCR$ ranked by \textbf{descending order};

\gls*{pr}: $Pr=\frac{TP}{TP+FP}$;

$Pr$ ranked by \textbf{descending order};

\gls*{f1}: $F1=2\times \frac{Pr.Re}{Pr+Re}$

$F1$ ranked by \textbf{descending order};

\gls*{r}:\\ $\gls*{r}=\frac{\overline{Re}+\overline{Sp}+\overline{FPR}+\overline{FNR}+\overline{WCR}+\overline{CCR}+\overline{F1}}{nMet}$;

\gls*{r} ranked by \textbf{ascending order};

$\overline{\gls*{arc}}$:\\
$\overline{\gls*{arc}}=\frac{Re+Sp+FPR+FNR+WCR+CCR+F1}{nMet}$;

$\overline{\gls*{arc}}$ ranked by \textbf{ascending order};

where $nMet$ is the number of metrics (8 in this case). 

The benchmark of the four algorithms (i.e. HSMDv1, HSMDv2, NeuroHSMDv1 and NeuroHSMDv2) and is required to ensure that the four algorithms produce comparative results to that of the original HSMDv1 results when tested against \gls*{cdnet2012} and \gls*{cdnet2014} datasets. Furthermore, the \gls*{opencl} calls the Intel Quartus, which performs several hardware optimisations that may include converting from floating-point to fixed-point representation \cite{Intel2017}, which might affect the accuracy of the \gls*{neurohsmd} algorithms during the synthesis step (one of the steps of the \gls*{opencl} design flow). 

\begin{table} 
\caption{FPGA resources usage.\strut} \label{tab:resources_usage_v1}
\resizebox{15.1cm}{!}{
\begin{tabular}{|l|l|l|l|l|l|}
\hline
\hline
\multicolumn{6}{|c|}{\Large{\textit{\textbf{Summary}}}} \\\hline\hline
\multicolumn{6}{|l|}{\Large\textbf{Info}} \\\hline
Project Names: & \multicolumn{5}{|l|}{} \\
 & \multicolumn{5}{|l|}{snn\_pc\_v1} \\
 & \multicolumn{5}{|l|}{snn\_pci\_v2} \\\hline
Target Family, Device, Board & \multicolumn{5}{|l|}{Stratix 10, 1SG280LU3F50E1VGS1, de10\_pro:s10\_sh2e1\_4Gx2} \\\hline
AOC Version & \multicolumn{5}{|l|}{19.1.0 Build 240} \\\hline
Quartus Version & \multicolumn{5}{|l|}{19.1.0 Build 240 Pro} \\\hline

\multicolumn{6}{|l|}{\Large{\textbf{Quartus Fit Clock Summary}}}\\\hline
Frequency (MHz) & \multicolumn{5}{|l|}{}\\
NeuroHSMDv1 kernel & \multicolumn{5}{|l|}{\cellcolor[HTML]{9B9B9B}306.25 (fmax)}\\
NeuroHSMDv2 kernel & \multicolumn{5}{|l|}{300 (Kernel fmax)} \\\hline\hline
\multicolumn{6}{|l|}{\Large{\textbf{Quartus Fit Resource Utilisation Summary}}} \\\hline
 & \textit{\glspl*{alm}} & \textbf{FFs} & \textbf{\glspl*{ram}} & \textbf{\gls*{dsp} blocks} & \textbf{\glspl*{mlab}}\\
NeuroHSMDv1 kernel & \cellcolor[HTML]{9B9B9B}488257.1 & 1060084 & 2484 & 1024 & \cellcolor[HTML]{9B9B9B}3871 \\
NeuroHSMDv2 kernel	& 486808.4	& \cellcolor[HTML]{9B9B9B}1034661 &	2486 &	1024 &	3933 \\\hline\hline
\multicolumn{6}{|c|}{\textbf{Kernel Resource Usage}} \\\hline
\multicolumn{1}{|c|}{} & \multicolumn{1}{c|}{\textbf{\glspl*{alut}}} & \multicolumn{1}{c|}{\textbf{FFs}} & \multicolumn{1}{c|}{\textbf{\glspl*{ram}}} & \multicolumn{1}{c|}{\textbf{\gls*{dsp} blocks}} & \multicolumn{1}{c|}{\textbf{\glspl*{mlab}}} \\\hline
NeuroHSMDv1 kernel & 488257.1 & 1060084 & \cellcolor[HTML]{9B9B9B}2484 & 1024 & \cellcolor[HTML]{9B9B9B}3871 \\
NeuroHSMDv2 kernel	& \cellcolor[HTML]{9B9B9B}486808.4	& \cellcolor[HTML]{9B9B9B}1034661 &	2486 &	1024 &	3933 \\\hline\hline
\multicolumn{6}{|c|}{\textbf{Global Interconnect}} \\\hline
NeuroHSMDv1 kernel & 10629 & 16485 & 61 & 0 & 0 \\
NeuroHSMDv2 kernel & 10629 & 16485 & 61 &	0 & 0 \\\hline
\multicolumn{6}{|c|}{\textbf{Board Interface}} \\\hline
NeuroHSMDv1 kernel & 13132 & 20030 & 112 & 0 & 0 \\
NeuroHSMDv2 kernel & 13132 & 20030 & 112 & 0 & 0 \\\hline
\multicolumn{6}{|c|}{\textbf{System description ROM}} \\\hline
NeuroHSMDv1 kernel & 2 & 71 & 2 & 0 & 0 \\
NeuroHSMDv2 kernel & 2 & 71 & 2 & 0 & 0 \\\hline 
\multicolumn{6}{|c|}{\textbf{Total}} \\\hline
NeuroHSMDv1 kernel & 567523 (30\%) & 907449 (24\%) & 2963 (25\%) & 976 (17\%) & \cellcolor[HTML]{9B9B9B} 3786 \\
NeuroHSMDv2 kernel  & \cellcolor[HTML]{9B9B9B} 554199 (30\%) & \cellcolor[HTML]{9B9B9B} 881379 (24\%) & 3043 (26\%) & 976 (17\%) & 3844 \\\hline\hline
\end{tabular}}
\end{table}

\section{Results} \label{Ch5:results}
The HSMDv1, HSMDv2, NeuroHSMDv1 and NeuroHSMDv2 were all tested on the same computer equipped with a quad-core Intel(R) Core(TM) i7-4770 \gls*{cpu} @ 3.40GHz, 16GB of DDR3 @ 1600 MHz and 1TB of \gls*{hdd}. 

The results section is divided into three parts. Namely, Section \ref{Ch4.1:resources_usage} shows the resources' usage to enable the comparison between the two kernels' complexity, the speed performance results are presented in Section \ref{Ch4.2:performance} and Section \ref{Ch4.3:benchmark} shows the benchmark results when tested against the \gls*{cdnet2012} and \gls*{cdnet2014} datasets.

\subsection{Resources Usage} \label{Ch4.1:resources_usage}
The resources' usage are provided in the report generated by the \textit{aoc} after the successful completion of the kernel compilation, which can take several hours (typically between 6h and 24h depending on the kernel complexity for the DE10pro). 
The resources' usage for the compilation of the NeuroHSMDv1 and NeuroHSMDv2 kernels is given in Table~\ref{tab:resources_usage_v1}. 

From the analysis of Table~\ref{tab:resources_usage_v1} it can be seen that the estimated resource utilisation is more pessimistic than the final resource utilisation, which is a direct consequence of the Intel Quartus's optimisations during the synthesis and routing phases and to ensure that the circuit fits in the \gls*{fpga} device. Nevertheless, it takes about 5 minutes to get the \textit{estimated resources usage} and between 6h to 24h to get the \textit{resources' utilisation summary}. Therefore, it is a good practice defining the coefficient \textbf{N} in the statement \textbf{\# PRAGMA UNROLL N} based on the \textit{estimated resources' usage}.

The NeuroHSMDv2 compared with the NeuroHSMDv1, consumes 1448.7 \glspl*{alm} less, 25423 FFs less, 2 \glspl*{ram} less and the same number of \glspl*{dsp}. Nevertheless, the NuroHSMDv2 kernel max frequency is 300MHz, while the NuroHSMDv1 kernel max frequency is 306.25 MHz. The NuroHSMDv2 enables saving of less than 1\% of resources and introduces an 2\% increase in latency.

The coefficient N should always be a multiple of $2^n$ to ensure the optimal use of \gls*{fpga} resources. For example, the resources' usage of $N=48$ is equivalent to $N=64$. Moreover, both \glspl*{ndk} had failed to compile when $N=32$ because there was not enough \glspl*{alut}. The design required more \glspl*{alut} than those available on the device, violating the compilation rules because the design would not fit on the device. 

Finally, the same neuronal parameters were used for all the neurons to reduce resource usage and the amount of data to be exchanged between the host \gls*{cpu} and the \gls*{fpga} device via the \gls*{pcie} bus. It is important to highlight that it is only possible to increase the size of the \gls*{snn} by reducing the complexity of the neuron model, and vice versa.

\subsection{Speed Performance} \label{Ch4.2:performance}
Table \ref{tab:cdnet2012_speed_results} displays the speed results obtained for the four algorithms tested against the \gls*{cdnet2012}.

\begin{table} \caption{\gls*{cdnet2012} speed result \strut} \label{tab:cdnet2012_speed_results}
\resizebox{14.5cm}{!}{
\begin{tabular}{|l|c|c|c|c|c|c|c|}
\hline
Category & number of & height & width & NeuroHSMDv2 & NeuroHSMDv1 & HSMDv2 & HSMDv1 \\ 
 & images & [pixels] & [pixels] & [fps] & [fps] & [fps] & [fps] \\ \hline
baseline/PETS2006 & 1199 & 576 & 720 & \cellcolor[HTML]{C0C0C0}23.66 & 20.45 & 8.40 & 9.73 \\ \hline
cameraJitter/badminton & 1149 & 480 & 720 & \cellcolor[HTML]{C0C0C0}28.33 & 25.48 & 10.01 & 11.50 \\ \hline
dynamicBackground/fall & 3999 & 480 & 720 & \cellcolor[HTML]{C0C0C0}27.28 & 25.01 & 9.87 & 11.08 \\ \hline
shadow/copyMachine & 3399 & 480 & 720 & \cellcolor[HTML]{C0C0C0}28.56 & 25.86 & 10.04 & 10.98 \\ \hline
dynamicBackground/fountain01 & 1183 & 288 & 432 & 55.17 & \cellcolor[HTML]{C0C0C0}58.30 & 27.40 & 29.99 \\ \hline
dynamicBackground/fountain02 & 1498 & 288 & 432 & 56.12 & \cellcolor[HTML]{C0C0C0}58.13 & 27.64 & 30.37 \\ \hline
intermittentObjectMotion/abandonedBox & 4499 & 288 & 432 & 55.93 & \cellcolor[HTML]{C0C0C0}58.45 & 26.88 & 29.89 \\ \hline
intermittentObjectMotion/tramstop & 3199 & 288 & 432 & 55.35 & \cellcolor[HTML]{C0C0C0}58.66 & 27.05 & 29.59 \\ \hline
thermal/park & 599 & 288 & 352 & 55.72 & \cellcolor[HTML]{C0C0C0}59.96 & 28.99 & 33.02 \\ \hline
shadow/peopleInShade & 1198 & 244 & 380 & 66.85 & \cellcolor[HTML]{C0C0C0}74.06 & 36.05 & 38.74 \\ \hline
baseline/highway & 1699 & 240 & 320 & 72.48 & \cellcolor[HTML]{C0C0C0}85.70 & 46.97 & 53.00 \\ \hline
baseline/office & 2049 & 240 & 360 & 69.49 & \cellcolor[HTML]{C0C0C0}78.86 & 40.24 & 46.71 \\ \hline
baseline/pedestrians & 1098 & 240 & 360 & 69.59 & \cellcolor[HTML]{C0C0C0}78.73 & 40.18 & 47.00 \\ \hline
cameraJitter/boulevard & 2499 & 240 & 352 & 68.96 & \cellcolor[HTML]{C0C0C0}78.81 & 41.21 & 46.98 \\ \hline
cameraJitter/sidewalk & 1199 & 240 & 352 & 69.04 & \cellcolor[HTML]{C0C0C0}78.54 & 39.49 & 47.32 \\ \hline
cameraJitter/traffic & 1569 & 240 & 320 & 72.17 & \cellcolor[HTML]{C0C0C0}85.29 & 43.95 & 51.40 \\ \hline
dynamicBackground/boats & 7998 & 240 & 320 & 71.86 & \cellcolor[HTML]{C0C0C0}84.29 & 45.33 & 51.70 \\ \hline
dynamicBackground/canoe & 1188 & 240 & 320 & 71.98 & \cellcolor[HTML]{C0C0C0}83.86 & 44.15 & 52.95 \\ \hline
dynamicBackground/overpass & 2999 & 240 & 320 & 71.98 & \cellcolor[HTML]{C0C0C0}84.56 & 43.69 & 49.31 \\ \hline
intermittentObjectMotion/parking & 2499 & 240 & 320 & 72.98 & \cellcolor[HTML]{C0C0C0}85.21 & 44.68 & 48.55 \\ \hline
intermittentObjectMotion/sofa & 2749 & 240 & 320 & 73.56 & \cellcolor[HTML]{C0C0C0}86.46 & 44.37 & 47.96 \\ \hline
intermittentObjectMotion/streetLight & 3199 & 240 & 320 & 72.01 & \cellcolor[HTML]{C0C0C0}84.95 & 44.31 & 47.76 \\ \hline
intermittentObjectMotion/winterDriveway & 2499 & 240 & 320 & 72.98 & \cellcolor[HTML]{C0C0C0}85.88 & 44.72 & 49.21 \\ \hline
shadow/backdoor & 1999 & 240 & 320 & 72.06 & \cellcolor[HTML]{C0C0C0}85.77 & 44.21 & 48.87 \\ \hline
shadow/bungalows & 1699 & 240 & 360 & 68.61 & \cellcolor[HTML]{C0C0C0}78.36 & 39.80 & 42.25 \\ \hline
shadow/busStation & 1249 & 240 & 360 & 68.79 & \cellcolor[HTML]{C0C0C0}78.78 & 39.85 & 42.91 \\ \hline
shadow/cubicle & 7399 & 240 & 352 & 70.87 & \cellcolor[HTML]{C0C0C0}80.47 & 41.07 & 44.52 \\ \hline
thermal/corridor & 5399 & 240 & 320 & 73.80 & \cellcolor[HTML]{C0C0C0}87.01 & 44.58 & 48.10 \\ \hline
thermal/diningRoom & 3699 & 240 & 320 & 73.38 & \cellcolor[HTML]{C0C0C0}86.60 & 44.38 & 47.82 \\ \hline
thermal/lakeSide & 6499 & 240 & 320 & 73.91 & \cellcolor[HTML]{C0C0C0}86.80 & 45.62 & 49.64 \\ \hline
thermal/library & 4899 & 240 & 320 & 74.83 & \cellcolor[HTML]{C0C0C0}87.05 & 44.40 & 49.36 \\ \hline
\end{tabular}}
\\
Best results are highlighted using grey.
\end{table}

From Table~\ref{tab:cdnet2012_speed_results} can be seen that both the NeuroHSMDv1 and NeuroHSMDv2 have performed better than the software versions (i.e. HSMDv1 and HSMDv2). It is also apparent that the NeuroHSMDv2 performs better in images with higher resolution (i.e. $720 \times 480$ and $720 \times 576$) while the NeuroHSMDv1 in lower resolutions (i.e below $720 \times 480$). Unlike in software, where it is consistent that the HSMDv2 is always faster than the HSMDv1 (non-opimised version), \gls*{fpga} optimisations require the utilisation of more resources which might increase latency. Therefore, the NeuroHSMDv2 is only more efficient for resolutions above $288\times 432$.

Overall, the NeuroHSMDv1 had an average frame rate of 71.50 \gls*{fps}, NeuroHSMDv2 63.20 \gls*{fps}, HSMDv1 40.26 \gls*{fps}, HSMDv2 36.11 \gls*{fps}. Finally, the average frame rate for processing images with the native resolution of $720 \times 480$ per algorithm is i) NeuroHSMDv2 28.06 \gls*{fps}, NeuroHSMDv1 25.45 \gls*{fps}, HSMDv2 11.19 \gls*{fps} and HSMDv1 9.97 \gls*{fps}.

The speed results obtained for the four algorithms when tested against the \gls*{cdnet2014} are depicted in Table~\ref{tab:cdnet2014_speed_results}

\begin{table}
\caption{\gls*{cdnet2014} speed results \strut} \label{tab:cdnet2014_speed_results}
\resizebox{14.5cm}{!}{
\begin{tabular}{|l|c|c|c|c|c|c|c|}
\hline
Category & number of & height & width & NeuroHSMDv2 & NeuroHSMDv1 & HSMDv2 & HSMDv1 \\ 
 & images & [pixels] & [pixels] & [fps] & [fps] & [fps] & [fps] \\ \hline
badWeather/blizzard & 6999 & 480 & 720 & \cellcolor[HTML]{C0C0C0}29.70 & 24.55 & 10.12 & 11.21 \\ \hline
badWeather/snowFall & 6499 & 480 & 720 & \cellcolor[HTML]{C0C0C0}29.61 & 24.31 & 10.16 & 11.11 \\ \hline
badWeather/wetSnow & 3499 & 540 & 720 & \cellcolor[HTML]{C0C0C0}26.54 & 21.76 & 8.93 & 10.19 \\ \hline
baseline/PETS2006 & 1199 & 576 & 720 & \cellcolor[HTML]{C0C0C0}25.04 & 20.36 & 8.52 & 9.53 \\ \hline
cameraJitter/badminton & 1149 & 480 & 720 & \cellcolor[HTML]{C0C0C0}28.58 & 25.07 & 9.80 & 11.00 \\ \hline
dynamicBackground/fall & 3999 & 480 & 720 & \cellcolor[HTML]{C0C0C0}27.48 & 24.75 & 13.90 & 10.96 \\ \hline
shadow/copyMachine & 3399 & 480 & 720 & \cellcolor[HTML]{C0C0C0}28.76 & 25.66 & 14.10 & 11.51 \\ \hline
turbulence/turbulence0 & 4999 & 480 & 720 & \cellcolor[HTML]{C0C0C0}28.59 & 25.07 & 14.26 & 11.62 \\ \hline
turbulence/turbulence1 & 3999 & 480 & 720 &  \cellcolor[HTML]{C0C0C0}28.24 & 25.21 & 14.28 &  11.35 \\ \hline
turbulence/turbulence3 & 2199 & 486 & 720 &  \cellcolor[HTML]{C0C0C0}28.72 & 25.15 & 14.07 & 11.49 \\ \hline
PTZ/continuousPan & 1699 & 480 & 704 & \cellcolor[HTML]{C0C0C0}28.49 & 23.91 & 9.88 & 10.85 \\ \hline
lowFramerate/tunnelExit\_0\_35fps & 3999 & 440 & 700 & \cellcolor[HTML]{C0C0C0}31.45 & 28.23 & 15.94 & 12.47 \\ \hline
nightVideos/fluidHighway & 1363 & 450 & 700 & \cellcolor[HTML]{C0C0C0}31.17 & 27.64 & 15.27 & 12.10 \\ \hline
turbulence/turbulence2 & 4499 & 315 & 645 &  \cellcolor[HTML]{C0C0C0}41.85 & 39.59 & 23.90 & 18.93 \\ \hline
lowFramerate/port\_0\_17fps & 2999 & 480 & 640 &  \cellcolor[HTML]{C0C0C0}31.35 & 28.17 & 15.75 & 12.39 \\ \hline
lowFramerate/tramCrossroad\_1fps & 899 & 350 & 640 & \cellcolor[HTML]{C0C0C0}39.51 & 36.73 & 21.37 & 16.71 \\ \hline
nightVideos/busyBoulvard & 2759 & 364 & 640 &  \cellcolor[HTML]{C0C0C0}38.86 & 36.00 & 20.90 & 16.43 \\ \hline
nightVideos/bridgeEntry & 2499 & 430 & 630 &  \cellcolor[HTML]{C0C0C0}34.52  & 31.85 & 17.90 & 14.20 \\ \hline
nightVideos/winterStreet & 1784 & 420 & 624 & \cellcolor[HTML]{C0C0C0}35.92 & 32.57 & 18.26 & 14.52 \\ \hline
nightVideos/streetCornerAtNight & 5199 & 245 & 595 & \cellcolor[HTML]{C0C0C0}52.86 & 51.56 & 32.83 & 25.35 \\ \hline
PTZ/twoPositionPTZCam & 2299 & 340 & 570 & \cellcolor[HTML]{C0C0C0}43.30 & 41.11 & 17.91 &     18.92 \\ \hline
PTZ/intermittentPan & 3499 & 368 & 560 &  \cellcolor[HTML]{C0C0C0}40.65 & 38.52 & 16.39 & 17.71 \\ \hline
badWeather/skating & 3899 & 360 & 540 &  \cellcolor[HTML]{C0C0C0}43.00 & 40.79 & 17.20 & 19.08 \\ \hline
nightVideos/tramStation & 2999 & 295 & 480 & \cellcolor[HTML]{C0C0C0}53.15 & 52.62 & 33.46 & 26.15 \\ \hline
dynamicBackground/fountain01 & 1183 & 288 & 432 & 55.38 & \cellcolor[HTML]{C0C0C0}57.06 & 37.95 & 30.16 \\ \hline
dynamicBackground/fountain02 & 1498 & 288 & 432 & 56.56 & \cellcolor[HTML]{C0C0C0}56.97 & 38.22 & 30.49 \\ \hline
intermittentObjectMotion/abandonedBox & 4499 & 288 & 432 & 55.64 & \cellcolor[HTML]{C0C0C0}58.03 & 38.19 & 30.26 \\ \hline
intermittentObjectMotion/tramstop & 3199 & 288 & 432 & 56.58 & \cellcolor[HTML]{C0C0C0}57.56 & 38.03 & 28.99 \\ \hline
shadow/peopleInShade & 1198 & 244 & 380 & 67.31 & \cellcolor[HTML]{C0C0C0}71.54 & 49.84 & 41.11 \\ \hline
baseline/office & 2049 & 240 & 360 & 71.72 & \cellcolor[HTML]{C0C0C0}75.86 & 37.85 & 45.39 \\ \hline
baseline/pedestrians & 1098 & 240 & 360 & 70.75 & \cellcolor[HTML]{C0C0C0}75.87 & 40.13 & 45.66 \\ \hline
shadow/bungalows & 1699 & 240 & 360 & 69.25 & \cellcolor[HTML]{C0C0C0}75.88 & 55.53 & 45.12 \\ \hline
shadow/busStation & 1249 & 240 & 360 & 69.51 & \cellcolor[HTML]{C0C0C0}74.76 & 55.41 & 46.24 \\ \hline
cameraJitter/boulevard & 2499 & 240 & 352 & 70.44 & \cellcolor[HTML]{C0C0C0}75.44 & 39.33 & 42.60 \\ \hline
cameraJitter/sidewalk & 1199 & 240 & 352 & 69.44 & \cellcolor[HTML]{C0C0C0}75.95 & 38.54 & 42.36 \\ \hline
shadow/cubicle & 7399 & 240 & 352 & 71.78 & \cellcolor[HTML]{C0C0C0}77.42 & 57.46 & 46.65 \\ \hline
thermal/park & 599 & 288 & 352 & 57.07 & \cellcolor[HTML]{C0C0C0}58.39 & 43.26 & 36.78 \\ \hline
PTZ/zoomInZoomOut & 1129 & 240 & 320 & 72.70 & \cellcolor[HTML]{C0C0C0}80.89 & 41.94 & 44.97 \\ \hline
baseline/highway & 1699 & 240 & 320 & 74.12 & \cellcolor[HTML]{C0C0C0}82.40 & 42.85 & 46.89 \\ \hline
cameraJitter/traffic & 1569 & 240 & 320 & 72.87 &  \cellcolor[HTML]{C0C0C0}81.54 & 40.97 & 45.95 \\ \hline
dynamicBackground/boats & 7998 & 240 & 320 & 73.14 & \cellcolor[HTML]{C0C0C0}82.14 & 60.35 & 47.99 \\ \hline
dynamicBackground/canoe & 1188 & 240 & 320 & 72.64 & \cellcolor[HTML]{C0C0C0}81.62 & 61.19 & 46.91 \\ \hline
dynamicBackground/overpass & 2999 & 240 & 320 & 72.85 & \cellcolor[HTML]{C0C0C0}82.26 & 61.72 & 50.00 \\ \hline
intermittentObjectMotion/parking & 2499 & 240 & 320 & 73.56 & \cellcolor[HTML]{C0C0C0}81.88 & 61.94 & 51.23 \\ \hline
intermittentObjectMotion/sofa & 2749 & 240 & 320 & 73.57 & \cellcolor[HTML]{C0C0C0}82.95 & 62.15 & 47.82 \\ \hline
intermittentObjectMotion/streetLight & 3199 & 240 & 320 & 72.77 & \cellcolor[HTML]{C0C0C0}82.48 & 62.27 & 48.13 \\ \hline
intermittentObjectMotion/winterDriveway & 2499 & 240 & 320 & 74.19 & \cellcolor[HTML]{C0C0C0}82.81 & 62.96 & 48.48 \\ \hline
lowFramerate/turnpike\_0\_5fps & 1499 & 240 & 320 &  73.10 & \cellcolor[HTML]{C0C0C0}82.38 & 60.99 &  46.52 \\ \hline
shadow/backdoor & 1999 & 240 & 320 & 73.73 & \cellcolor[HTML]{C0C0C0}83.84 & 62.82 & 51.77 \\ \hline
thermal/corridor & 5399 & 240 & 320 & 74.86 &  \cellcolor[HTML]{C0C0C0}83.65 & 62.48 & 51.11 \\ \hline
thermal/diningRoom & 3699 & 240 & 320 & 74.42 & \cellcolor[HTML]{C0C0C0}83.86 & 62.34 & 50.98 \\ \hline
thermal/lakeSide & 6499 & 240 & 320 & 74.93 & \cellcolor[HTML]{C0C0C0}83.60 & 63.23 & 51.68 \\ \hline
thermal/library & 4899 & 240 & 320 & 75.79 & \cellcolor[HTML]{C0C0C0}83.72 & 62.17 & 47.23 \\ \hline
\end{tabular}}
\\
Best results are highlighted using grey.
\end{table}

\begin{figure*}[htb!]
\centering
\includegraphics[width=1\textwidth]{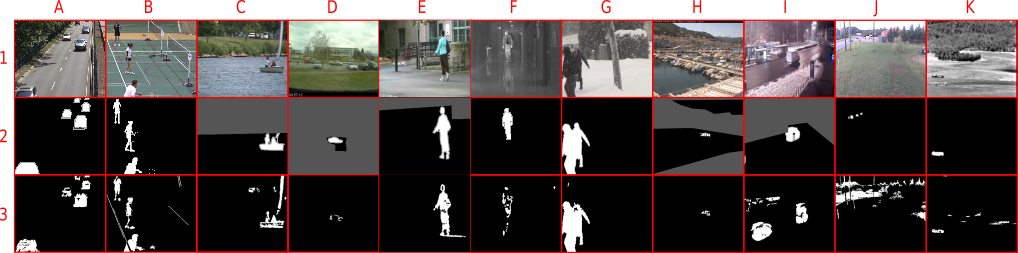}
\caption{Results obtained for each of the eleven of the five categories (columns A to F) are common to both \gls*{cdnet2012} and \gls*{cdnet2014} datasets, while the remaining six categories (columns G to K) are only available on \gls*{cdnet2014} dataset. Column A: baseline; B: camera jitter; C: dynamic background; D: dynamic object motion; E: shadow, F: thermal, G: bad weather, H: low frame rate; I: night videos, J: PTZ and K: turbulence. Row 1: RGB image; 2: ground-truth; and 3: \gls*{neurohsmd} binarised. The raw images, shown in the first row, demonstrate the scenarios that can be found in both datasets. The corresponding ground truth images, presented in the second row, show the 5 labels, namely, i) static [greyscale value 0], ii) shadow [greyscale value 50], iii) non-\gls*{roi} [greyscale value 85], iv) unknown [greyscale value 170] and v) moving [greyscale value 255]. The corresponding binarised images generated by the \gls*{neurohsmd} are shown in the third row.} \label{fig:CDnet2012_output}
\end{figure*}

Table~\ref{tab:cdnet2014_speed_results} shows that the NeuroHSMDv1 and NeuroHSMDv2 have performed better than the software versions (i.e. HSMDv1 and HSMDv2) when tested against the \gls*{cdnet2014} dataset. Once again, the NeuroHSMDv2 performs
\hfill\\
The results for each of the eleven categories shared by both \gls*{cdnet2012} and \gls*{cdnet2014} are shown in Figure~\ref{fig:CDnet2012_output}.
better in images with higher resolution (i.e. equal or higher than $480 \times 295$) while the NeuroHSMDv1 in lower resolutions
(i.e below $480 \times 295$). Again, the NeuroHSMDv2 is only more efficient for resolutions above $288\times 432$ because of the complexity and latency introduced by the optimisation circuit.

Overall, the NeuroHSMDv1 has an average frame rate of 43.51 \gls*{fps}, NeuroHSMDv2 39.94 \gls*{fps}, HSMDv2 29.30 \gls*{fps}, and HSMDv1 25.18 \gls*{fps}.  It is essential to highlight that the \gls*{cdnet2014} has more categories and image sequences, leading to different frame rates for the \gls*{cdnet2012} and \gls*{cdnet2014} datasets.  

The average frame rate for processing images with the native resolution of $720 \times 480$ per algorithm was i) NeuroHSMDv2 28.71 \gls*{fps}, NeuroHSMDv1 24.95 \gls*{fps}, HSMDv2 12.37 \gls*{fps} and HSMDv1 11.25 \gls*{fps}. These results are in line with the results obtained for the 4 algorithms when tested against the \gls*{cdnet2012} dataset. Although HSMDv1 and HSMDv2 have archived near-real time performance, it is also clear that \glspl*{cpu} are not ideal for accelerating \glspl*{snn}. Furthermore, to achieve such performance on the \gls*{cpu} all the non-essential applications were closed to ensure that all the memory and computational resources were free to compute the HSMDv1 and HSMDv2 with maximum efficiency. Furthermore, the parallelisation process takes place in the \gls{fpga}, where the device kernel is defined in its internal logic. Data is exchanged between the device and host \glspl*{sdram} through buffers. The \gls*{fpga} implementation contains 16 parallel circuits, which were configured using the IOCL. In this work, we did not explore learning rules (such as \gls*{stdp}, Hebbian, Oja, etc.) as these would require additional \gls*{fpga} resources.

\subsection{Benchmark} \label{Ch4.3:benchmark}
Table~\ref{tab:cdnet2012_ranks} shows the results obtained after testing the 4 methods against the \gls*{cdnet2012} ground-truth images using the scripts provided by Nil Goyette et al. \cite{Goyette2012}.

\begin{table} \caption{\gls*{cdnet2012} Overall ranks \strut} \label{tab:cdnet2012_ranks}
\resizebox{14.5cm}{!}{
\begin{tabular}{|l|>{\columncolor[gray]{0.8}}c|c|c|c|c|c|c|c|c|}
\hline
Method & $\overline{\gls*{arc}}$ $\downarrow$ & \gls*{re} $\uparrow$ & \gls*{sp} $\uparrow$ & \gls*{fpr} $\downarrow$ & \gls*{fnr} $\downarrow$ & \gls*{wcr} $\downarrow$ & \gls*{ccr} $\uparrow$ & \gls*{f1} $\uparrow$ & \gls*{pr} $\uparrow$ \\ \hline\hline
\rowcolor{maroon!10}
HSMDv1 & \cellcolor[HTML]{D0C0C0}2.8 & 0.52 & 0.994 & 0.006 & 0.23 & 0.024 & 0.976 & 0.77 & 0.62 \\ \hline
\rowcolor{maroon!10}
HSMDv2 & \cellcolor[HTML]{D0C0C0}2.8 & 0.52 & 0.994 & 0.006 & 0.23 & 0.024 & 0.976 & 0.77 & 0.62 \\ \hline
\rowcolor{maroon!10}
NeuroHSMDv1 & \cellcolor[HTML]{D0C0C0}2.8 & 0.52 & 0.994 & 0.006 & 0.23 & 0.024 & 0.976 & 0.77 & 0.62 \\ \hline
\rowcolor{maroon!10}
NeuroHSMDv2 & \cellcolor[HTML]{D0C0C0}2.8 & 0.52 & 0.994 & 0.006 & 0.23 & 0.024 & 0.976 & 0.77 & 0.62 \\ \hline
GSOC \cite{Machado2021}&3.5&0.54&0.993&0.007&0.25&0.024&0.976&0.75&0.63\\\hline
MOG2 \cite{Machado2021}&3.8&0.37&0.995&0.004&0.24&0.026&0.974&0.76&0.50\\\hline
GMG \cite{Machado2021}&3.9&0.20&0.998&0.002&0.21&0.033&0.967&0.79&0.32\\\hline
KNN \cite{Machado2021}&4.3&0.39&0.995&0.005&0.26&0.025&0.975&0.74&0.51\\\hline
MOG \cite{Machado2021}&4.5&0.32&0.996&0.004&0.26&0.030&0.970&0.74&0.44\\\hline
CNT \cite{Machado2021}&6.1&0.73&0.927&0.073&0.71&0.081&0.919&0.29&0.41\\\hline
LSBP\cite{Machado2021}&7.3&0.57&0.90&0.096&0.80&0.109&0.891&0.20&0.29\\\hline
\end{tabular}}
\\
$\uparrow$: the highest score is the best. $\downarrow$: the lowest result is the best. \\
All the 4 methods were ranked first because no changes were made to the customised \gls*{snn}.
$\gls*{re}$ stands for Recall, $\gls*{sp}$ Specificity, \gls*{fpr} False Positive Rate, \gls*{fnr} False Negative Rate, \gls*{wcr} Wrong Classifications Rate, \gls*{ccr} Correct Classifications Rate, \gls*{pr} Precision, \gls*{f1} F-score and $\overline{\gls*{arc}}$ Average Ranking across all Categories.
\end{table}

From the results shown in Table~\ref{tab:cdnet2012_ranks} it is possible to infer that the results obtained with the four methods are the same because all the methods were ranked in first place with the same values per metric. Indexing all the algorithms in the first place was expected because the speed optimisation in version 2 of the \gls*{neurohsmd} and \gls*{hsmd} should not interfere with the model dynamics. Furthermore, the HSMDv1, HSMDv2, NeuroHSMDv1 and NeuroHSMDv2  showed poor performance in both dynamic backgrounds and low frame rate conditions, indicating that the spiking neuron model is not effective in accurately distinguishing the type of motion. This is likely due to the fact that in the vertebrate retina, only ganglion cells are spiking cells and the distinction between the main object and shadows is performed by other non-spiking cells. However, the integration of the \gls*{gsoc} algorithm and the \gls*{snn} in a new approach has significantly improved the accuracy of the \gls*{gsoc} algorithm by mimicking the basic functionality of \gls*{oms-gc}.

Table~\ref{tab:cdnet2014_ranks} depicts the results obtained after testing 4 methods against the \gls*{cdnet2014} ground-truth images using the scripts provided by Nil Goyette et al. \cite{Goyette2012}.

\begin{table} \caption{\gls*{cdnet2014} Overall ranks \strut} \label{tab:cdnet2014_ranks}
\resizebox{14.5cm}{!}{
\begin{tabular}{|l|>{\columncolor[gray]{0.8}}c|c|c|c|c|c|c|c|c|}
\hline
Method & $\overline{\gls*{arc}}$ $\downarrow$ & \gls*{re} $\uparrow$ & \gls*{sp} $\uparrow$ & \gls*{fpr} $\downarrow$ & \gls*{fnr} $\downarrow$ & \gls*{wcr} $\downarrow$ & \gls*{ccr} $\uparrow$ & \gls*{f1} $\uparrow$ & \gls*{pr} $\uparrow$ \\ \hline\hline
\rowcolor{maroon!10}
HSMDv1 & \cellcolor[HTML]{D0C0C0}2.9 & 0.55 & 0.993 & 0.007 & 0.35 & 0.018 & 0.982 & 0.65 & 0.60\\ \hline
\rowcolor{maroon!10}
HSMDv2 & \cellcolor[HTML]{D0C0C0}2.9 & 0.55 & 0.993 & 0.007 & 0.35 & 0.018 & 0.982 & 0.65 & 0.60 \\ \hline
\rowcolor{maroon!10}
NeuroHSMDv1 & \cellcolor[HTML]{D0C0C0}2.9 & 0.55 & 0.993 & 0.007 & 0.35 & 0.018 & 0.982 & 0.65 & 0.60 \\ \hline
\rowcolor{maroon!10}
NeuroHSMDv2 & \cellcolor[HTML]{D0C0C0}2.9 & 0.55 & 0.993 & 0.007 & 0.35 & 0.018 & 0.982 & 0.65 & 0.60 \\ \hline
GSOC \cite{Machado2021}&3.0&0.40&0.995&0.005&0.38&0.017&0.983&0.62&0.48\\\hline
KNN \cite{Machado2021}&3.5&0.34&0.996&0.004&0.32&0.019&0.981&0.68&0.45\\\hline
GMG\cite{Machado2021}&4.3&0.24&0.997&0.003&0.36&0.022&0.978&0.64&0.35\\\hline
MOG \cite{Machado2021}&4.4&0.58&0.991&0.009&0.39&0.019&0.981&0.61&0.60\\\hline
MOG2 \cite{Machado2021}&4.5&0.39&0.994&0.006&0.42&0.018&0.982&0.58&0.47\\\hline
LSBP \cite{Machado2021}&6.5&0.58&0.945&0.055&0.79&0.064&0.936&0.21&0.31\\\hline
CNT&7.0 \cite{Machado2021}&0.72&0.930&0.070&0.80&0.075&0.925&0.20&0.32\\\hline
\end{tabular}}
\\
$\uparrow$: the highest score is the best. 
$\downarrow$: the lowest result is the best. \\
All the 4 methods were ranked first because no changes were made to the customised \gls*{snn}.
$\gls*{re}$ stands for Recall, $\gls*{sp}$ Specificity, \gls*{fpr} False Positive Rate, \gls*{fnr} False Negative Rate, \gls*{wcr} Wrong Classifications Rate, \gls*{ccr} Correct Classifications Rate, \gls*{pr} Precision, \gls*{f1} F-score and $\overline{\gls*{arc}}$ Average Ranking across all Categories.
\end{table}

From the results shown in Table~\ref{tab:cdnet2014_ranks} it is possible to infer that the results obtained with the four methods are the same because the four methods were ranked first with the same values per metric. These results are important because it is possible to infer that there has been no degradation in accuracy as a consequence of the hardware acceleration. Again, the HSMDv1, HSMDv2, NeuroHSMDv1, and NeuroHSMDv2 demonstrated subpar results in dynamic backgrounds and low frame rate situations, demonstrating that the spiking neuron model is not capable of accurately identifying the type of motion. This is likely a result of only ganglion cells being spiking cells in the vertebrate retina, where the distinction between the main object and shadows is handled by non-spiking cells. However, combining the \gls*{gsoc} algorithm and the \gls*{snn} in a novel approach has significantly enhanced the accuracy of the \gls*{gsoc} algorithm by replicating the basic operations of \gls*{oms-gc}.

\section{Conclusions and Future Work}\label{Ch6:conclusion}

Two bio-inspired \gls*{neurohsmd} have been proposed to accelerate the \gls*{hsmd} algorithm \cite{Machado2021}. The NeuroHSMDv1 and NeuroHSMDv2 (speed optimisation) were tested against the \gls*{cdnet2012} and \gls*{cdnet2014} datasets. The NeuroHSMDv1 has lower latency when processing images with resolutions equal to or greater than $480 \times 295$. The NeuronHSMDv2 (speed optimisation) has a lower latency when processing images with resolutions smaller than $480 \times 295$. Two \gls*{hsmd} versions were used (the original HSMDv1 algorithm \cite{Machado2021} and HSMDv2 with speed optimisation) for ensuring a fair comparison between the software and hardware implementations.
The HSMDv1, HSMDv2, NeuroHSMDv1 and NeuroHSMDv2 where all tested on the same computer equipped with a quad-core Intel(R) Core(TM) i7-4770 \gls*{cpu} @ 3.40GHz, 16GB of DDR3 @ 1600 MHz and 1TB of \gls*{hdd}. The average frame rate for processing images with the native resolution of $720 \times 480$ per algorithm was: 1) \textbf{\gls*{cdnet2012}:} i) NeuroHSMDv2 28.06 \gls*{fps}, NeuroHSMDv1 25.45 \gls*{fps}, HSMDv2 11.19 \gls*{fps} and HSMDv1 9.97 \gls*{fps}; and 2) \textbf{\gls*{cdnet2014}:}
i) NeuroHSMDv2 28.71 \gls*{fps}, NeuroHSMDv1 24.95 \gls*{fps}, HSMDv2 12.37 \gls*{fps} and HSMDv1 11.25 \gls*{fps}.\\
The four methods were also tested against the ground-truth images available in the \gls*{cdnet2012} and \gls*{cdnet2014} datasets using the eight metrics, were used to assess and compare the quality of the \gls*{hsmd} algorithm. The four methods obtained the same values for all the metrics and were all ranked first. The first place acquired by the four methods is an indication that there was no degradation with the hardware acceleration.\\ 
Finally, the \gls*{neurohsmd} is the first Neuromorphic \gls*{snn} accelerator capable of accelerating thousands of spiking neuron models in parallel using \gls*{opencl}. Moreover, the proposed method can be used by non-engineering background users for accelerating \glspl*{snn} in different heterogenous platforms using \gls*{opencl}. It is also possible to conclude that the results on the \gls*{fpga} are the same as the results obtained in the \gls*{cpu} implementation, meaning that the target \gls*{fpga} offers sufficient IEEE754 single-precision \gls*{dsp} blocks to accelerate the \gls*{snn} kernel without degradation of the results.

Future work includes optimising the \gls*{hsmd} algorithm to detect and track motion in challenging scenarios (e.g. low frame rate, dynamic background and camera jitter) and optimise those \glspl*{snn} to run in affordable lower-end \glspl*{fpga}. The implementation and evaluation of more complex retinal cells (such as direction sensitive and predictive cells) using \glspl*{snn} and target lower-end and affordable \gls*{fpga} devices is also planned.\\

\bibliographystyle{ieeetr}
\bibliography{references}
\end{document}